\documentclass{article}

\usepackage[final,nonatbib]{neurips_2020}

\usepackage[utf8]{inputenc} 
\usepackage[T1]{fontenc}    
\usepackage{subfigure}
\usepackage{hyperref} 
\usepackage{url} 
\usepackage{booktabs} 
\usepackage{amsfonts} 
\usepackage{nicefrac} 
\usepackage{microtype} 
\usepackage{wrapfig}
\usepackage{multirow}
\usepackage{graphicx}
\usepackage{url}

\usepackage{multicol}

\usepackage{stackengine}
\newcommand\barbelow[1]{\stackunder[1.2pt]{$#1$}{\rule{.8ex}{.075ex}}}

\def\ourH{HyperCube}
\def\our{HyperCube-{Interval}}
\def\ourmultiline{\begin{tabular}{c} HyperCube \\ (Interval) \end{tabular}}

\title{\ourH{}: Implicit Field Representations of Voxelized 3D Models}

\author{%
  Magdalena Proszewska\\
 Jagiellonian University, \\
 Kraków, Poland\\
 \texttt{magdalena.proszewska@student.uj.edu.pl } \\
  \And
  Marcin Mazur \\
 Jagiellonian University, \\
 Kraków, Poland\\
 \texttt{marcin.mazur@uj.edu.pl } \\
  \AND
  Tomasz Trzciński  \\
  Warsaw University of \\
  Technology, \\
  Warsaw, Poland\\
  Tooploox\\
  \texttt{tomasz.trzcinski@pw.edu.pl } \\
  \And
  Przemys\l{}aw Spurek\\
  Jagiellonian University, \\
  Kraków, Poland\\
  \texttt{przemyslaw.spurek@uj.edu.pl} \\
}

\usepackage{amsmath,amsfonts,amssymb,amsthm,dsfont,stmaryrd}
\usepackage[utf8]{inputenc}
\usepackage{hyperref}

\usepackage{graphicx}

\usepackage{amssymb}

\usepackage{color}

\def\X{\mathcal{X}}

\def\R{\mathbb{R}}

\def\U{\mathcal{U}}
\def\X{\mathcal{X}}
\def\y{y}

\def\our{HyperFlow}

\theoremstyle{remark}

\begin{document}

\maketitle

\begin{abstract}

Recently introduced implicit field representations offer an effective way of generating 3D object shapes. They leverage implicit decoder trained to take a 3D point coordinate concatenated with a shape encoding and to output a value which indicates whether the point is outside the shape or not. Although this approach enables efficient rendering of visually plausible objects, it has two significant limitations. First, it is based on a single neural network dedicated for all objects from a training set which results in a cumbersome training procedure and its application in real life. More importantly, the implicit decoder takes only points sampled within voxels (and not the entire voxels) which yields problems at the classification boundaries and results in empty spaces within the rendered mesh.

To solve the above limitations, we introduce a new \ourH{} architecture based on interval arithmetic network, that enables direct processing of 3D voxels, trained using a hypernetwork paradigm to enforce model convergence. 
Instead of processing individual 3D samples from within a voxel, our approach allows to input the entire voxel (3D cube) represented with its convex hull coordinates, while the target network constructed by a hypernet assigns it to an inside or outside category. 
As a result our \ourH{} model outperforms the competing approaches both in terms of training and inference efficiency, as well as the final mesh quality. 

\end{abstract}

\section{Introduction}

Recently introduced implicit field representations of 3D objects offer high quality generations of 3D shapes~\cite{chen2019learning}. This method relies on an implicit decoder, called IM-NET, trained to take a 3D point coordinate concatenated with a shape encoding and to assign it a value indicating whether it belongs inside or outside of a given shape. 
Using this formulation, a shape is constructed out of points assigned to the inside class and it is typically rendered via the iso-surface extraction method such as Marching Cubes.


\begin{figure}[t!]
	\centering
	\begin{tabular}{@{}cc@{}c@{}c@{}c@{}c@{}}
	\rotatebox{90}{ \quad input} &
	\includegraphics[scale=0.09,trim={2cm 0cm 2cm 0cm},clip]{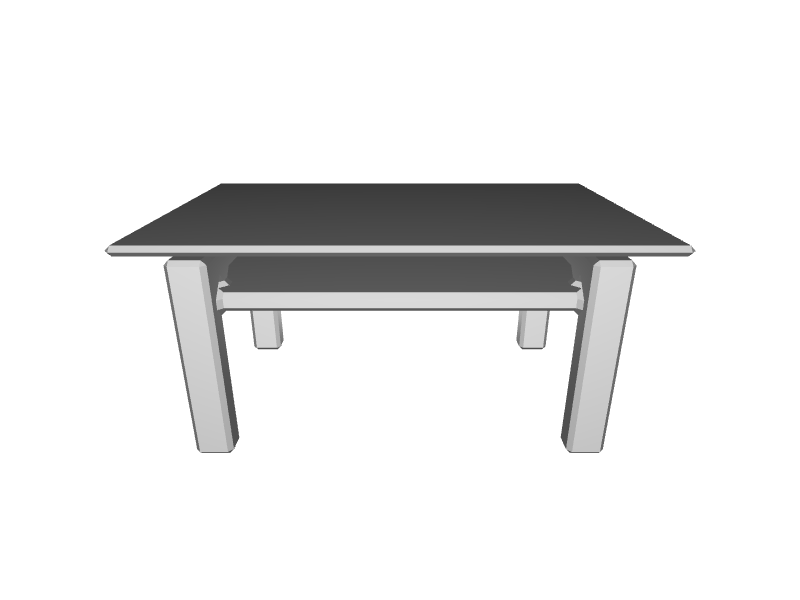}
	\includegraphics[scale=0.09,trim={2cm 0cm 2cm 0cm},clip]{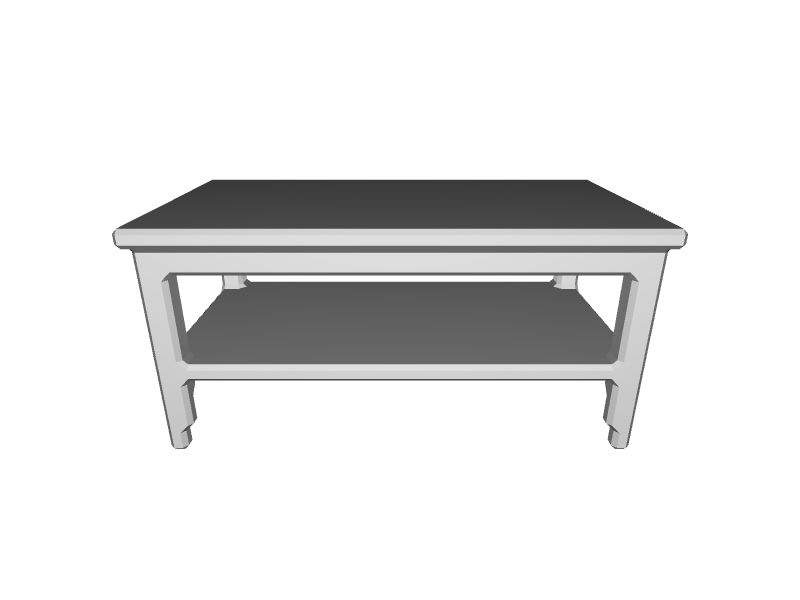}
    \includegraphics[scale=0.09,trim={2cm 0cm 2cm 0cm},clip]{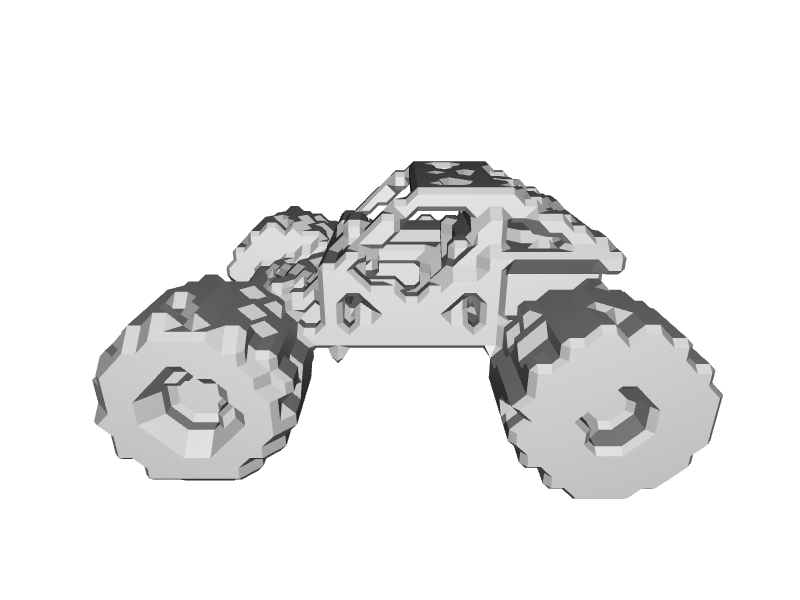}
    \includegraphics[scale=0.09,trim={2cm 0cm 2cm 0cm},clip]{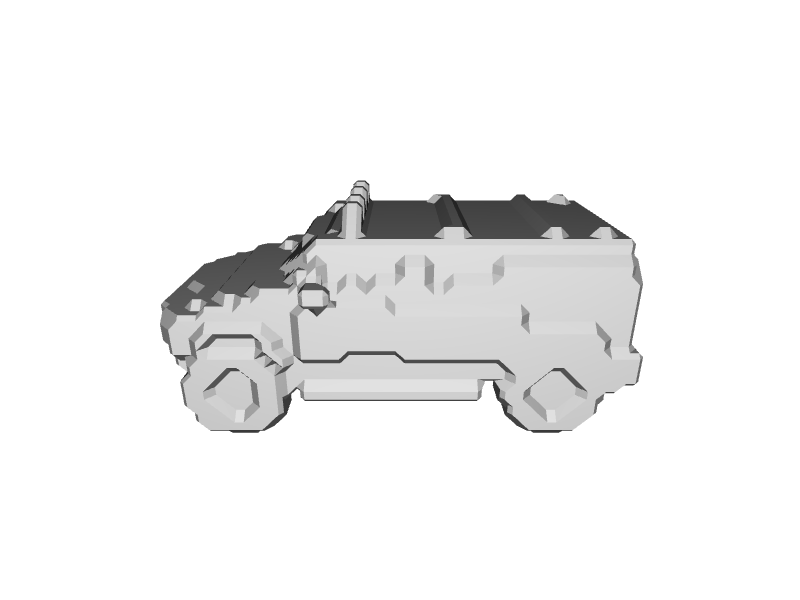}
    \includegraphics[scale=0.09,trim={2cm 0cm 2cm 0cm},clip]{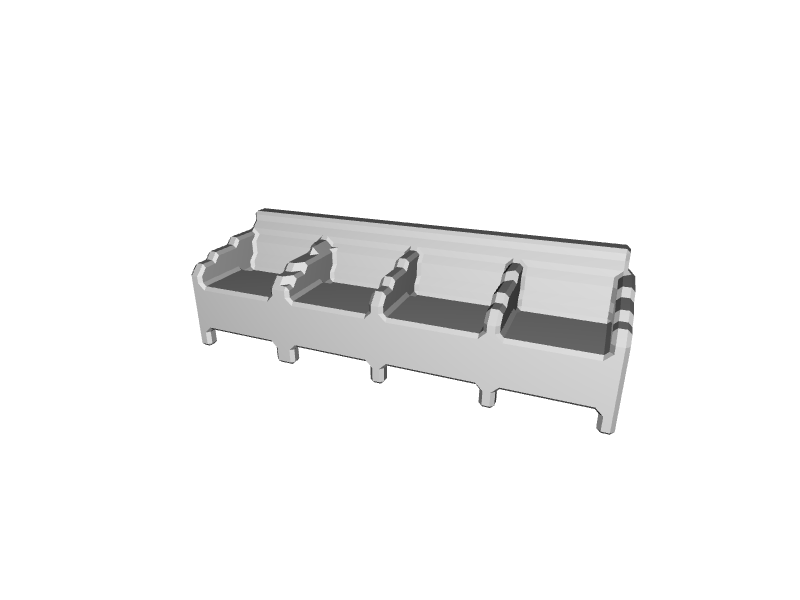}
    \includegraphics[scale=0.09,trim={2cm 0cm 2cm 0cm},clip]{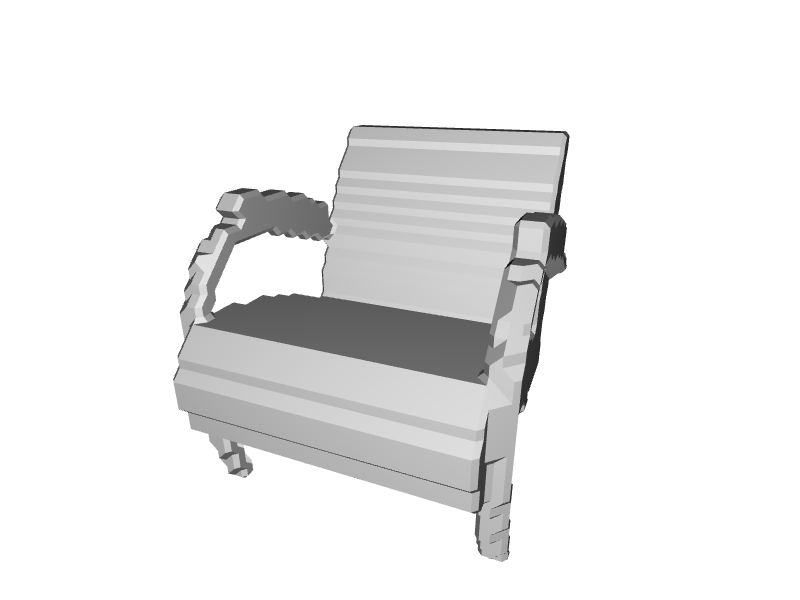}\\
    \rotatebox{90}{ \quad IM-NET} &
    \includegraphics[scale=0.09,trim={2cm 0cm 2cm 0cm},clip]{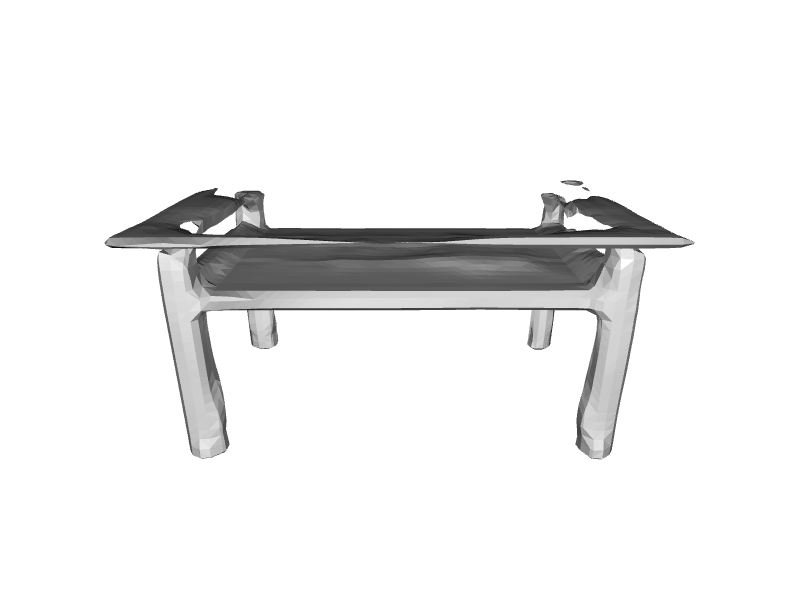}
    \includegraphics[scale=0.09,trim={2cm 0cm 2cm 0cm},clip]{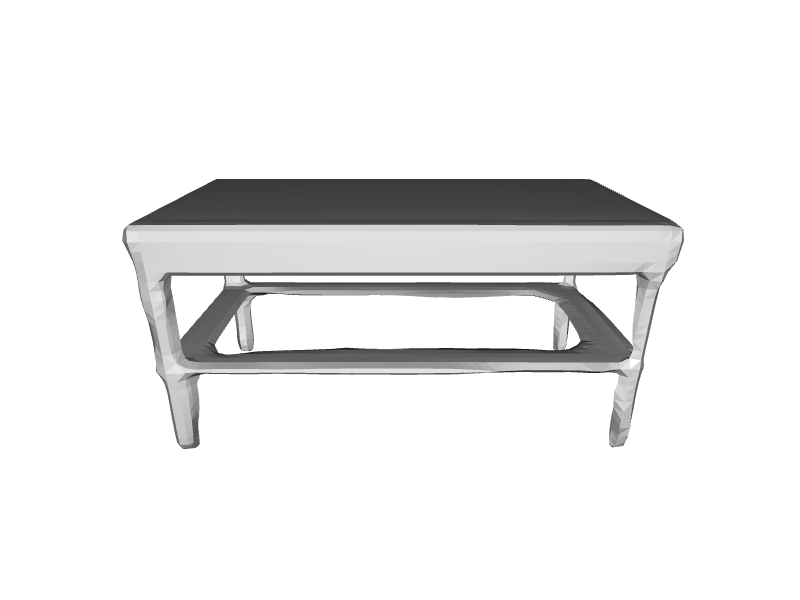}
    \includegraphics[scale=0.09,trim={2cm 0cm 2cm 0cm},clip]{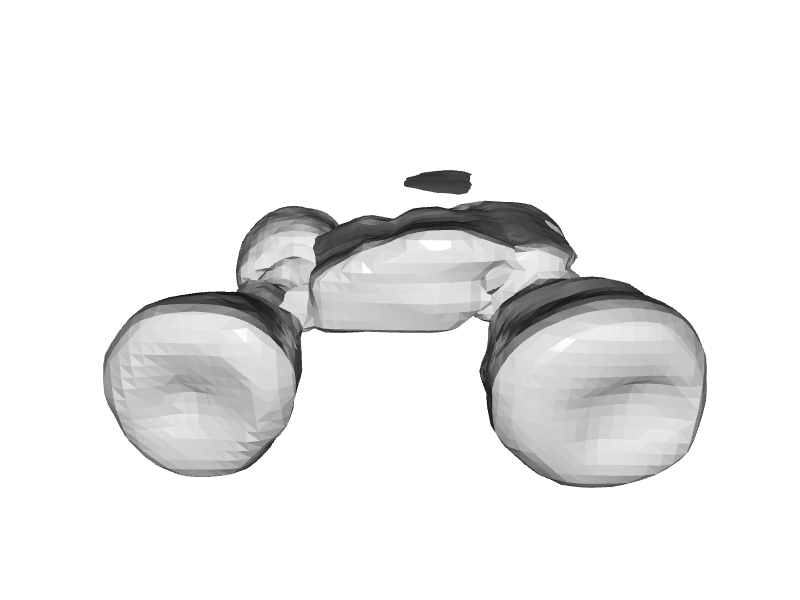}
    \includegraphics[scale=0.09,trim={2cm 0cm 2cm 0cm},clip]{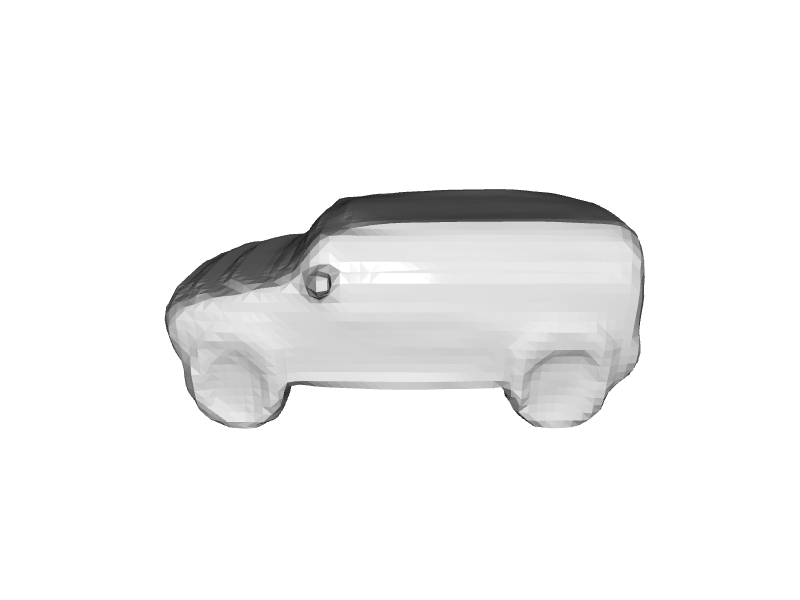}
    \includegraphics[scale=0.09,trim={2cm 0cm 2cm 0cm},clip]{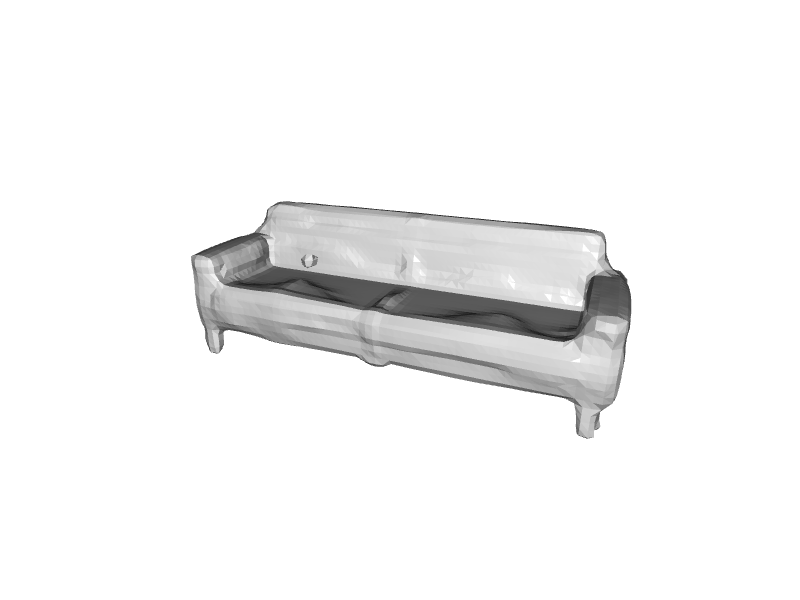}
    \includegraphics[scale=0.09,trim={2cm 0cm 2cm 0cm},clip]{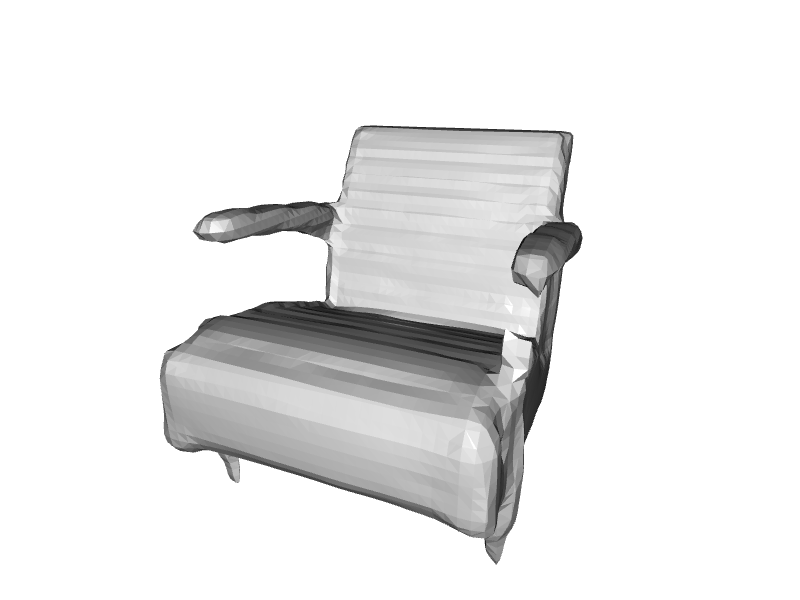}\\
    \rotatebox{90}{ \quad \ourH } &
    \includegraphics[scale=0.09,trim={2cm 0cm 2cm 0cm},clip]{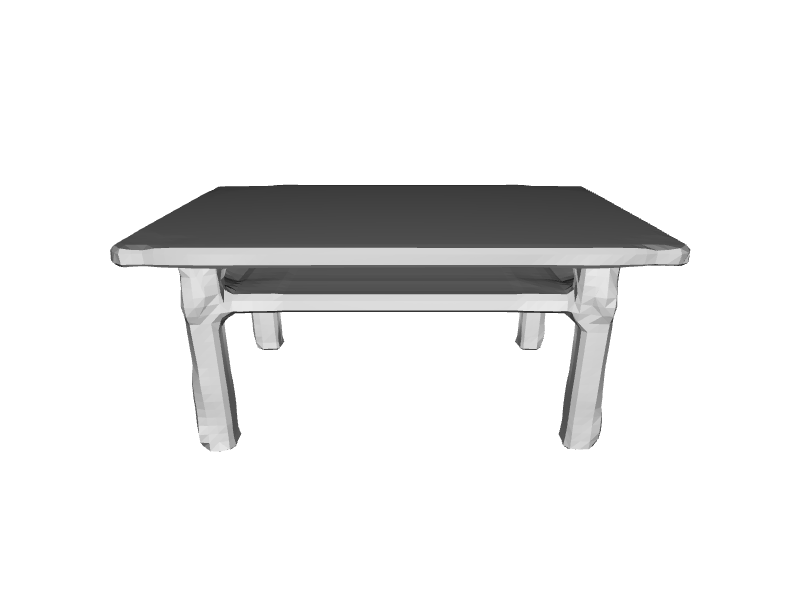}
    \includegraphics[scale=0.09,trim={2cm 0cm 2cm 0cm},clip]{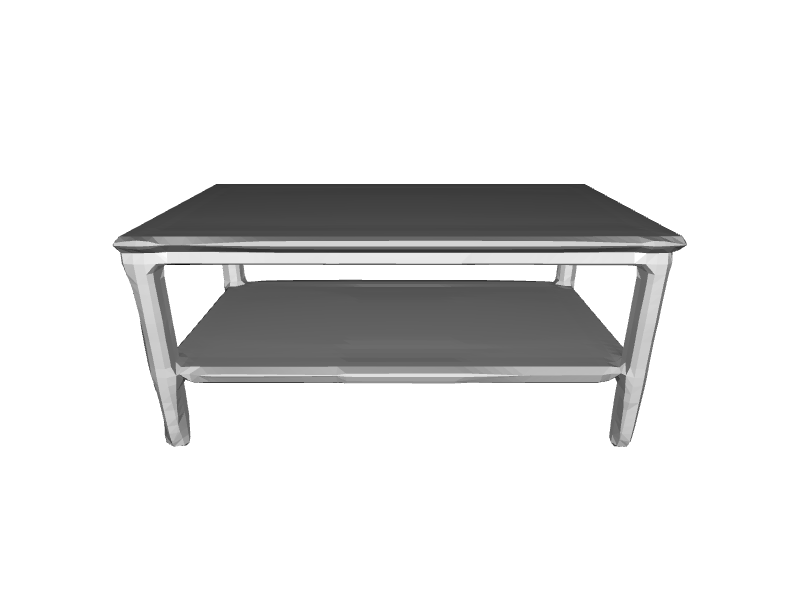}
    \includegraphics[scale=0.09,trim={2cm 0cm 2cm 0cm},clip]{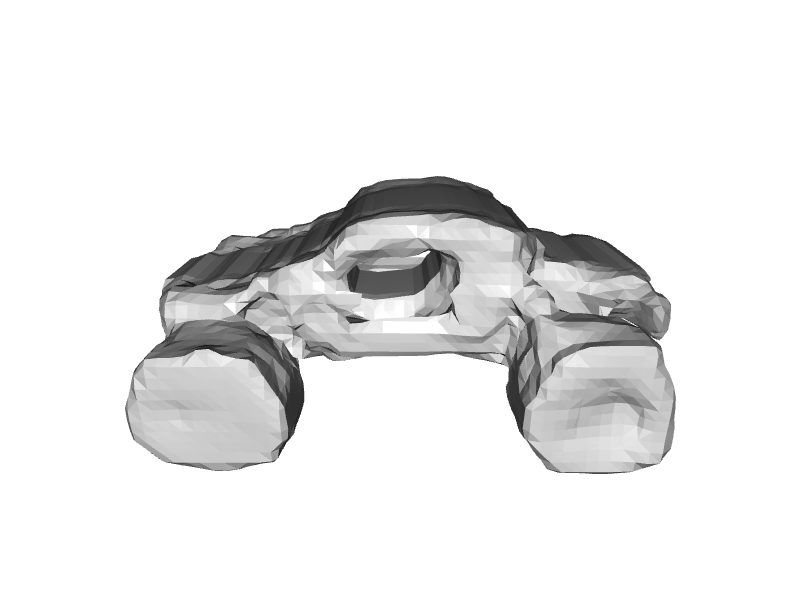}
    \includegraphics[scale=0.09,trim={2cm 0cm 2cm 0cm},clip]{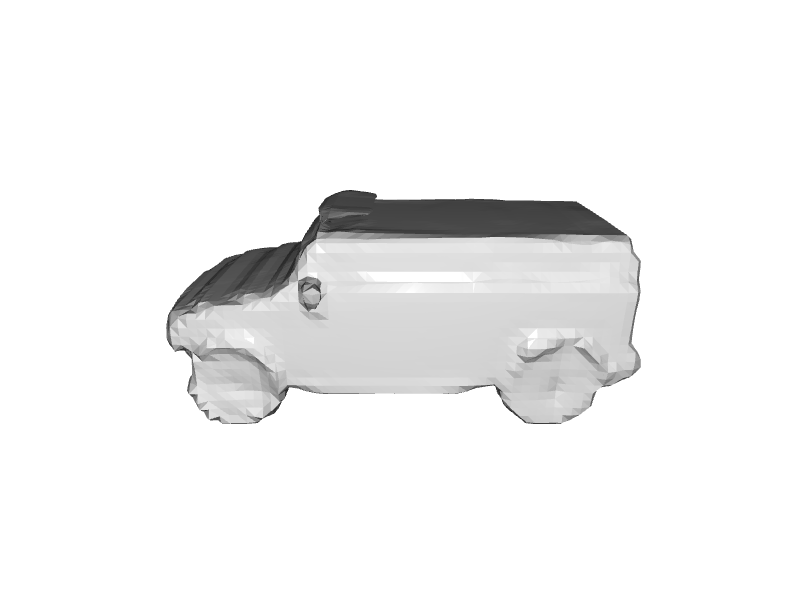}
    \includegraphics[scale=0.09,trim={2cm 0cm 2cm 0cm},clip]{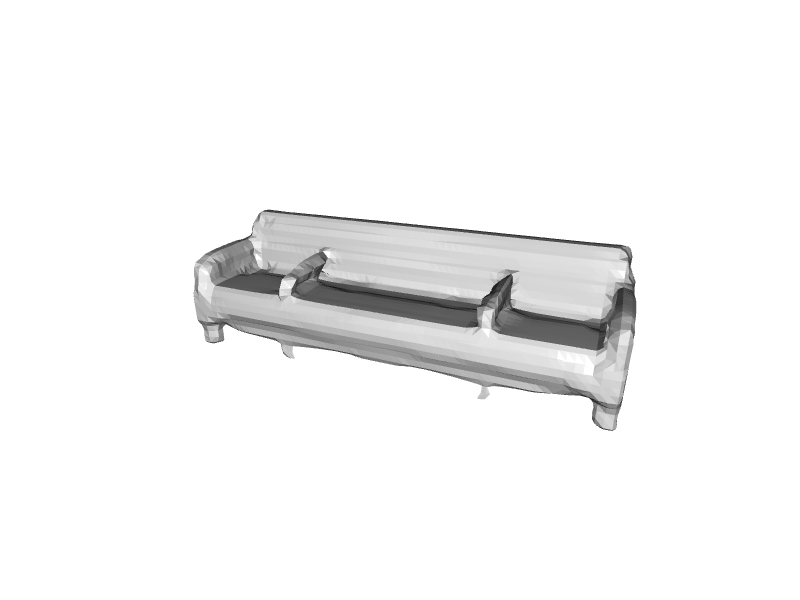}
	\includegraphics[scale=0.09,trim={2cm 0cm 2cm 0cm},clip]{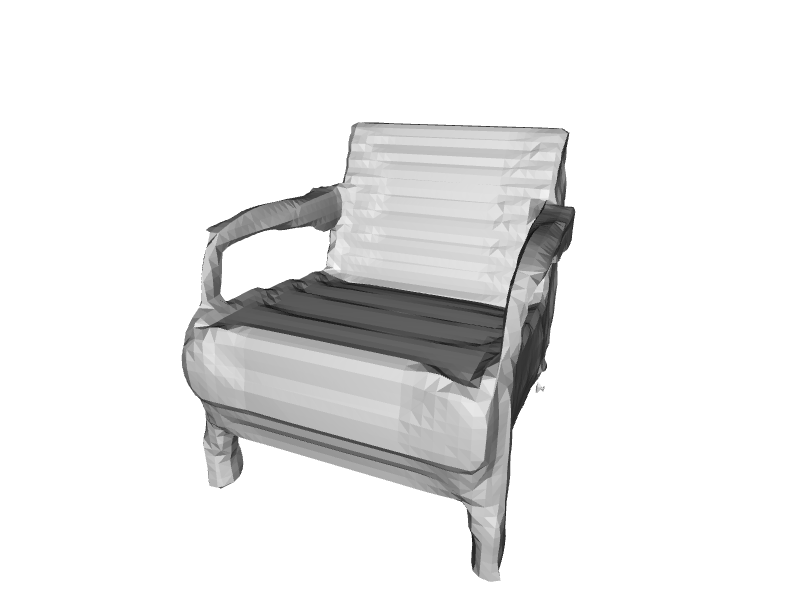}\\
		\end{tabular}
	\caption{Comparison between our \ourH{} approach and the state-of-the-art implicit field method based on the IM-NET~\cite{chen2019learning}. IM-NET takes as an input 3D points sampled within voxels, while \ourH{} leverages an interval arithmetic to process the entire voxels. As a result, \ourH{} offers a high quality 3D object rendering without missing important parts of object close to the implicit decoder's decision boundary, as done by IM-NET.}
\label{fig:whols}
\end{figure}

The IM-NET architecture has several advantages over a standard convolutional model. First of all, it can produce outputs of various resolutions, including those not observed in the training. Furthermore, IM-NET learns shape boundaries instead of voxel distributions over the volume, which results in surfaces of a higher quality. 

On the other hand, however, IM-NET has some limitations. First of all, the point coordinates processed by the model are concatenated with the shape embedding and to reconstruct an object the model needs to possess the knowledge about all objects present in the entire dataset. Therefore IM-NET architecture is hard to train on many different classes. Moreover, the implicit decoder processes only points sampled from within a voxel, instead of the entire voxels. This, in turn, yields problems at the classification boundaries at object edges and gives severe rendering artifacts, see Fig.~\ref{fig:whols}. 

In this paper, we address the above limitations by introducing a novel approach to 3D object representation based on implicit fields called \ourH{}\footnote{We make our implementation available at \url{https://github.com/mproszewska/HyperCube}}. Contrary to the baseline IM-NET model, our approach leverages a hyper-network architecture to produce weights of a target implicit decoder, based on the input feature vector defining a voxel. This target decoder assigns an {\it inside} or {\it outside} of a shape label to each processed voxel. Such architecture is more compact than IM-NET and therefore much faster to train, while it does not need to know the distribution of all objects in the training dataset to obtain object reconstructions. Furthermore, its design allows a flexible adjustment of the target network processing feature vectors. This enables us to input the entire voxels into the model leveraging interval arithmetic and the IntervalNet architecture~\cite{gowal2018effectiveness,morawiecki2019fast} and leading to the inception of our~\our{} model. 
The~\our{} architecture takes as an input relatively small 3D cubes (hence the name), instead of 3D points samples within the voxels. Therefore, it does not produce empty space in reconstructed mesh representation, as done by the IM-NET and visualized in Fig.~\ref{fig:whols}.


To summarize, our contributions are the following:
\begin{itemize}
    \item We introduce a new \ourH{} architecture for representing voxelized 3D models based on implicit field representation.
    \item We incorporate a hypernetwork paradigm into our architecture which leads to significant simplification of the resulting model and reduces training time. 
    \item Our approach offers unprecedented flexibility of integrating various target network models, and we show on the example of the IntervalNet how this can be leveraged to work with the entire voxels, and not their sampled versions, which significantly improves the quality of generated 3D models.
\end{itemize}

\section{Related works}
\label{gen_inst}

3D  objects can be represented using different approaches including voxel grids~\cite{choy20163d,girdhar2016learning,liao2018deep,wu2016learning}, octrees \cite{hane2017hierarchical,riegler2017octnetfusion,tatarchenko2017octree,wang2018adaptive}, multi-view images \cite{arsalan2017synthesizing,lin2018learning,su2015multi}, point clouds \cite{achlioptas2018learning,fan2017point,qi2017pointnet,qi2017pointnet++,yang2018foldingnet}, geometry
images \cite{sinha2016deep,sinha2017surfnet}, deformable meshes \cite{girdhar2016learning,sinha2017surfnet,wang2018pixel2mesh,yang2018foldingnet},
and part-based structural graphs \cite{li2017grass,zhu2018scores}.

All the above representations are discreet which hinders their application in real-life scenarios. Therefore recent works introduced 3D representations modeled as a continuous function~\cite{dupont2021generative}. In such a case the implicit occupancy~\cite{chen2019learning,mescheder2019occupancy,peng2020convolutional}, distance field~\cite{michalkiewicz2019implicit,park2019deepsdf} and surface parametrization~\cite{yang2019pointflow,spurek2020hypernetwork,spurek2020hyperflow,cai2020learning} models use a neural network to represent a 3D object. These methods do not use discretization (e.g., fixed number of voxels, points, or vertices), but represent shapes in a continuous manner and handle complicated shape topologies. 

In \cite{mescheder2019occupancy} the authors propose that the occupancy networks implicitly represent the 3D surface as a continuous decision boundary of a deep neural network classifier. Thus, the occupancy networks produce continuous representation instead of a discrete one, and render realistic high-resolution meshes.
In \cite{chen2019learning}, which is the most relevant baseline to our approach, the authors propose an implicit decoder (IM-NET). Such a model uses a binary classifier that takes a point coordinate concatenated with a feature vector encoding a shape and outputs a value that indicates inside or outside label. 
The above-mentioned implicit approaches are limited to rather simple geometries of single objects and do not scale to more complicated or large-scale
scenes. In \cite{peng2020convolutional}, the authors propose Convolutional
Occupancy Networks, dedicated for the representations of 3D scenes. The authors use convolutional encoders and implicit decoders. In general, all above method uses conditioning mechanism. In our paper, we use the hyper network, which gives a more flexible model which can be trained faster.

In \cite{park2019deepsdf} the authors introduce DeepSDF representation of 3D objects that produce high-quality meshes. DeepSDF represents a shape surface by a continuous volumetric field. Points represent the distance to the surface boundary, and the
sign indicates whether the region is inside or outside. Such representation implicitly encodes a shape’s border as a classification boundary.

In \cite{spurek2020hypernetwork,spurek2020hyperflow} the authors propose HyperCloud model that uses a hyper network to output weights of a generative network to create 3D point clouds, instead of generating a fixed size reconstruction. 
One neural network is trained to produce a continuous representation of an object.
In \cite{yang2019pointflow} authors propose to use a conditioning mechanism to produce a flow model which transfers gaussian noise into a 3D object. 

Our solution can be interpreted as a generalization of all the above methods. We use data reprehension approach described in IM-NET, but extend it to follow the hypernetwork paradigm used in HyperCloud. As a consequence, we take the best of both wordls and obtain a reconstruction quality of the IM-NET, while reducing the training and inference time as in the case of a HyperCloud. 


\begin{figure}[!t] \centering
	\subfigure[IM-NET]{\label{fig:architectur_1}
	\includegraphics[width=0.38\textwidth]{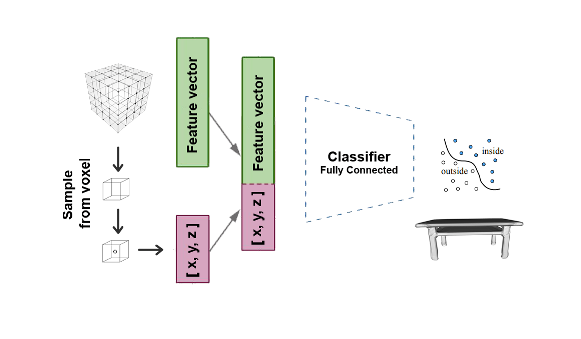}} 
	\! \! \! 
	\subfigure[\ourH{}]{\label{fig:architectur_2}\includegraphics[width=0.28\textwidth]{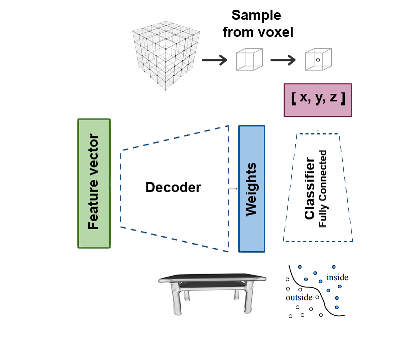}}
	\ 
	\subfigure[\our{}]{\label{fig:architectur_3}\includegraphics[width=0.28\textwidth]{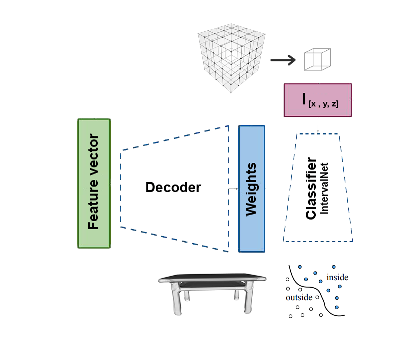}}
	\caption{Comparison of the network architectures: (a) IM-NET, (b) \ourH{} and (c) \our{}. (a) IM-NET uses a binary classifier that takes a point coordinate $(x, y, z)$ concatenated with a feature vector encoding a shape and outputs a value which indicates whether the point is outside the shape or not. IM-NET is a single neural network dedicated to all object from the training dataset. 
	(b) \ourH{} takes feature vectors and produce a new network, dubbed {\it target network} which classifies points sampled from voxel into one of two categories: inside or outside. For object reconstruction, only a target network run on a 3D cube is needed, which makes this solution significantly faster. (c) \our{} extends \ourH{} idea by incorporating interval arithmetic which in turn allows to process the entire voxel by an IntervalNet, instead of samples of voxels, as done in the IM-NET.}
	\label{fig:architectur} 
\end{figure}

\section{Hyper Implicit Field}

In this section, we present the \ourH{} model designed to produce a continuous representation of Implicit Fields. More precisely, for each 3D object represented by voxels, we would like to create a neural network that classifies elements from $\R^3$ to the inside or outside classes. 
We define voxel representation as a unit cube $[0,1] \times [0,1] \times [0,1]$ divided into small cubes with a given resolution $n \in \mathbb{N}$:

\begin{multline*}
V_n = \left\{ [\barbelow x_{i}, \bar x_{i}]  \times [\barbelow y_{i}, \bar y_{i}] \times [\barbelow z_{i}, \bar z_{i}]  \mbox{ where } \barbelow x_{i}, \barbelow y_{i}, \barbelow z_{i} = 0, \frac{1}{n}, \frac{2}{n}, \ldots, \frac{n-1}{n}\right.  \\
\left. \mbox{ and } \bar x_{i} = \barbelow x_{i} + \frac{1}{n}, \bar y_{i} = \barbelow y_{i} + \frac{1}{n}, \bar z_{i} = \barbelow z_{i} + \frac{1}{n} \right\}.
\end{multline*}

During the training, we use binary labels for each element of the voxel grid:
$
\X = \left\{ I_i \right\}_{i=1}^{n^3} , 
\y \subset \{ 0, 1 \}^{n^3},
$
where
$I_i = [\barbelow x_{i}, \bar x_{i}]  \times [\barbelow y_{i}, \bar y_{i}] \times [\barbelow z_{i}, \bar z_{i}] \in V_n$.
Although the labels are assigned to the voxels defined as 3D cubes, a classical neural networks are not able to process them in the raw continuous format. Therefore, such methods use points uniformly sampled from a voxel:
$
X = \left\{ (x_i,y_i,z_i): (x_i,y_i,z_i) \sim  \U_{I_i} \right\}_{i=1}^{n^3} , 
y \subset \{ 0, 1 \}^{n^3},
$ where $\U_{I_i}$ is uniform distribution in 3D cube $I_i$.
In the paper, we present a method that works directly with voxels.

We assume that we have feature vectors for each element from the training set. Our goal is to build a~continuous representation of 3D objects. More precisely we have to model function $f \colon \R^3 \to \{0,1\} $, which takes coordinate of 3D point and returns an inside/outside label.  Reconstruction of the object is produced by labeling all elements from the voxel grid using the Marching Cubes algorithm.

In this section, we first describe 
a general hyper network architecture used in our model. We follow up with the description of our \ourH{} method, which uses hyper networks to efficiently process 3D points. Finally,  we introduce interval arithmetic, which allows us to propagate 3D cubes instead of point sampled from voxels, and we show how we can use incorporate is within our approach \our{}.

\subsection{Hypernetwork}

Hyper networks, introduced in \cite{ha2016hypernetworks}, are defined as neural models that generate weights for target network solving a specific task. 
The authors aim to reduce trainable parameters by designing a hyper-network with a smaller number of parameters than the target network. 



In the context of 3D point clouds, various methods make use of a hyper network to produce a continuous representation of objects~\cite{spurek2020hypernetwork,spurek2021hyperpocket,spurek2021modeling}. HyperCloud~\cite{spurek2020hypernetwork} proposes a generative autoencoder-based model that relies on a decoder to produce a vector of weights $\theta$ of a target network $\mathcal{T}_\theta\colon \mathbb{R}^3\to \mathbb{R}^3$. The target network is designed to transform a prior into target objects (reconstructed data). In practice, we have one neural network architecture that uses different weights for each 3D object.

\begin{figure}[h!]
	\centering
	\begin{tabular}{@{}cc@{}c@{}c@{}c@{}}
    \rotatebox{90}{ \quad \ourH } &
    \includegraphics[scale=0.09,trim={2cm 1.5cm 2cm 1.5cm},clip]{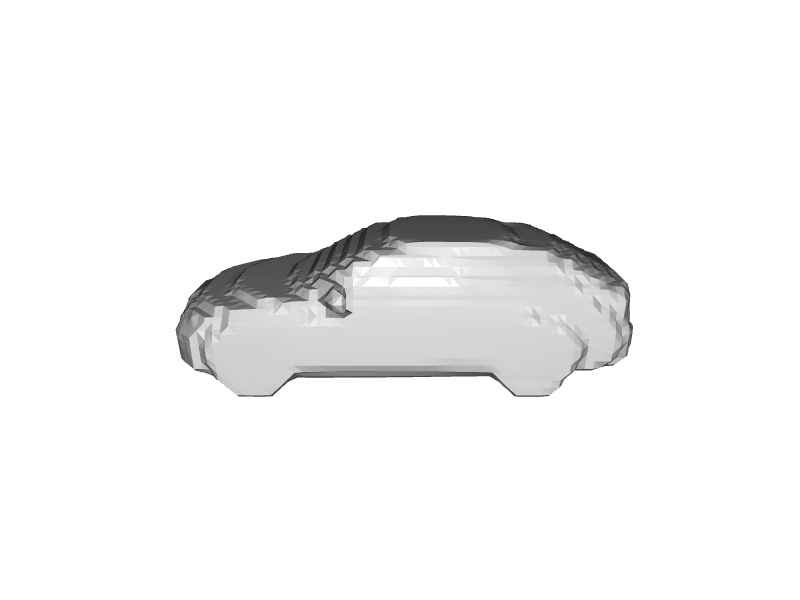}
    \includegraphics[scale=0.09,trim={2cm 1.5cm 2cm 1.5cm},clip]{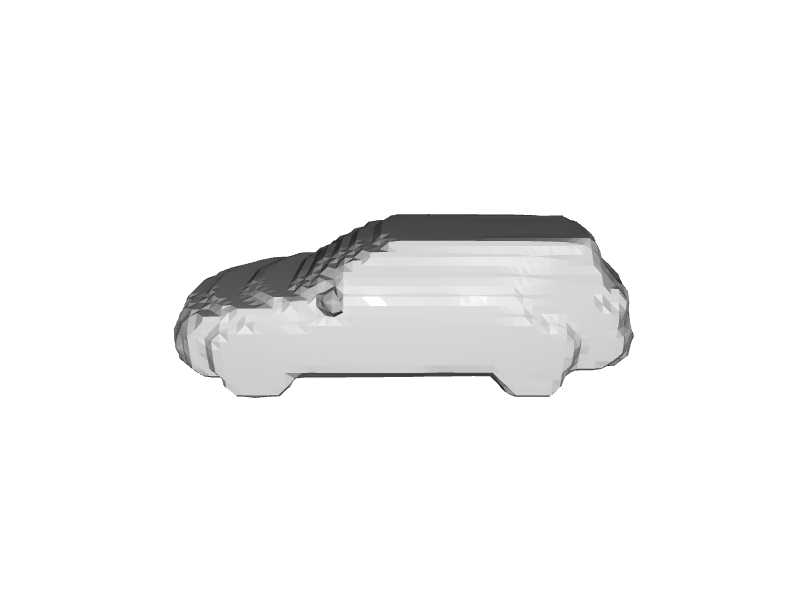}
    \includegraphics[scale=0.09,trim={2cm 1.5cm 2cm 1.5cm},clip]{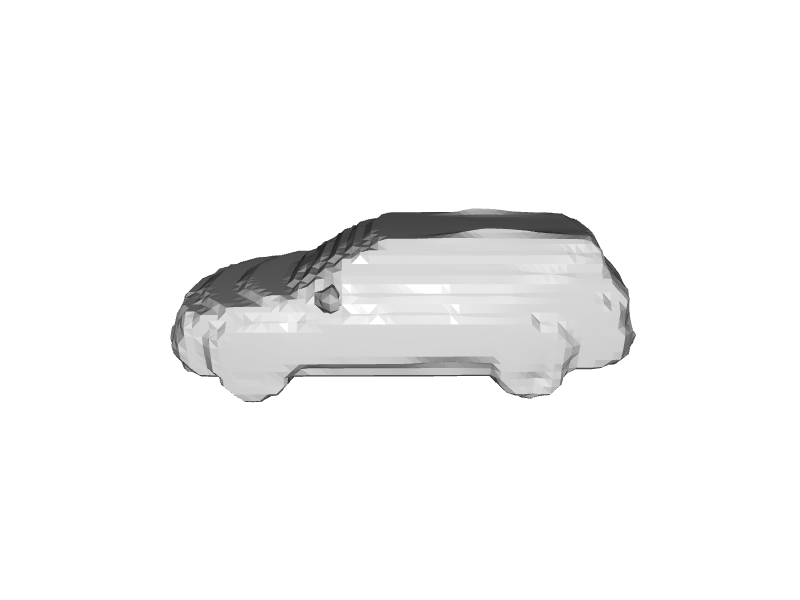}
	\includegraphics[scale=0.09,trim={2cm 1.5cm 2cm 1.5cm},clip]{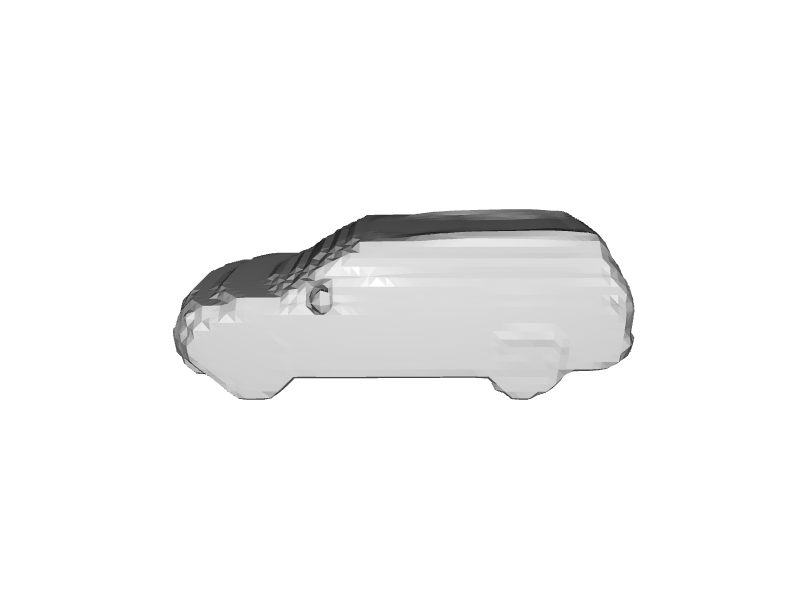}
	\includegraphics[scale=0.09,trim={2cm 1.5cm 2cm 1.5cm},clip]{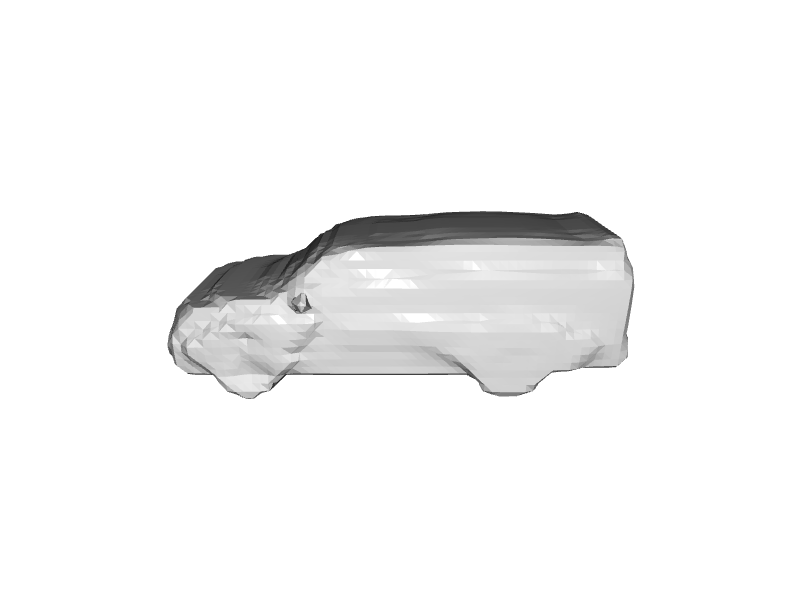}
	\includegraphics[scale=0.09,trim={2cm 1.5cm 2cm 1.5cm},clip]{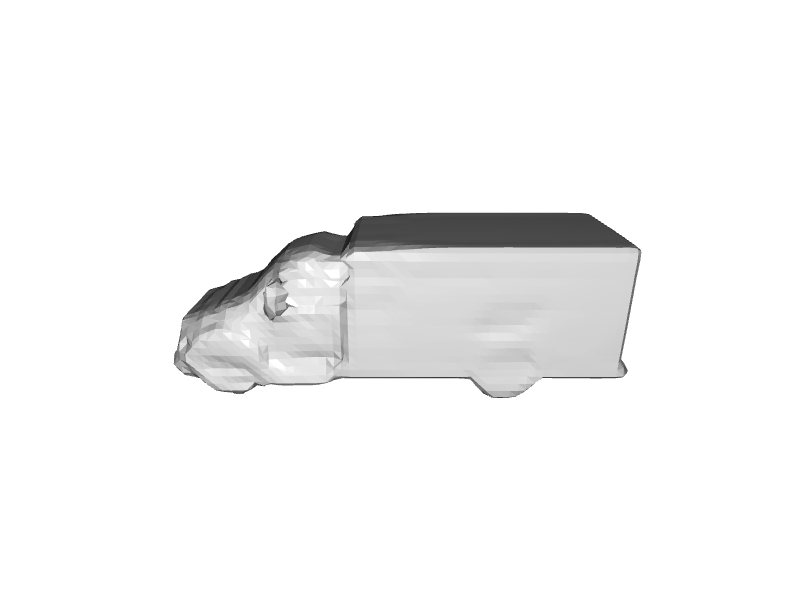}\\
	\rotatebox{90}{ \quad \ourH } &
	 \includegraphics[scale=0.09,trim={2cm 1.5cm 2cm 1.5cm},clip]{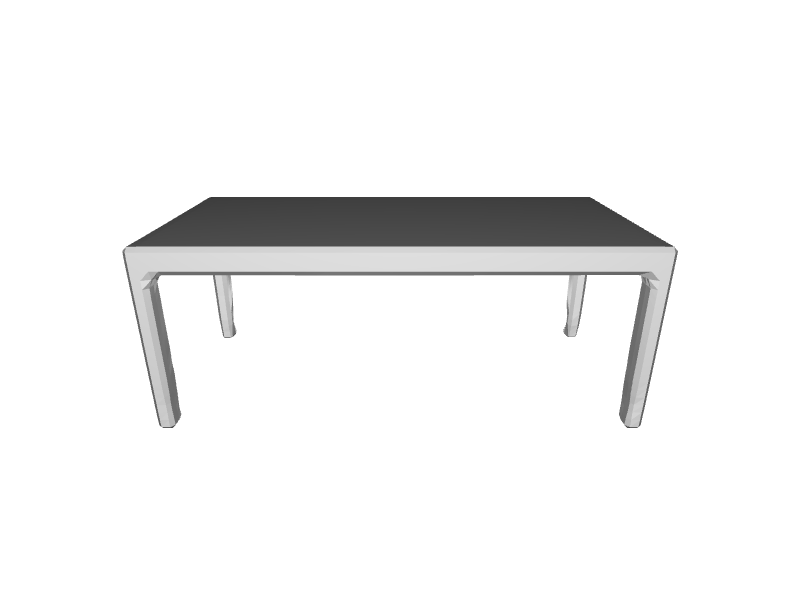}
    \includegraphics[scale=0.09,trim={2cm 1.5cm 2cm 1.5cm},clip]{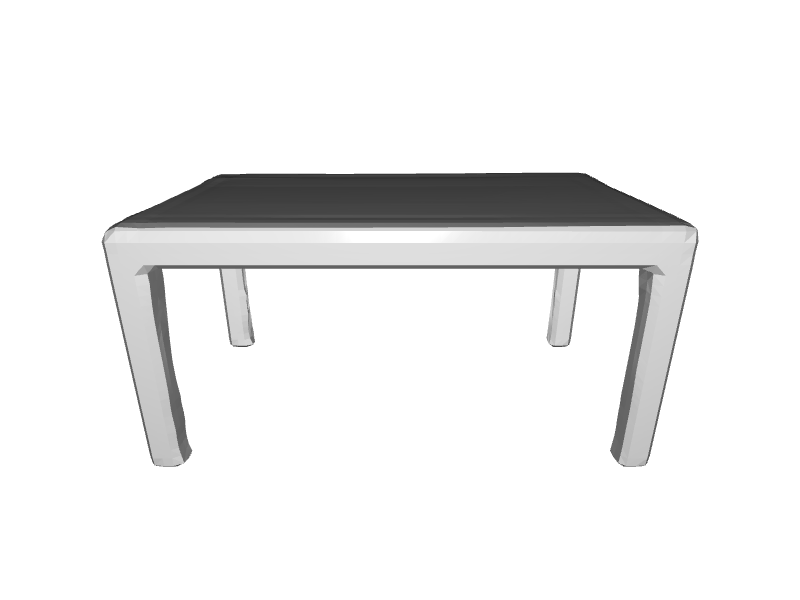}
    \includegraphics[scale=0.09,trim={2cm 1.5cm 2cm 1.5cm},clip]{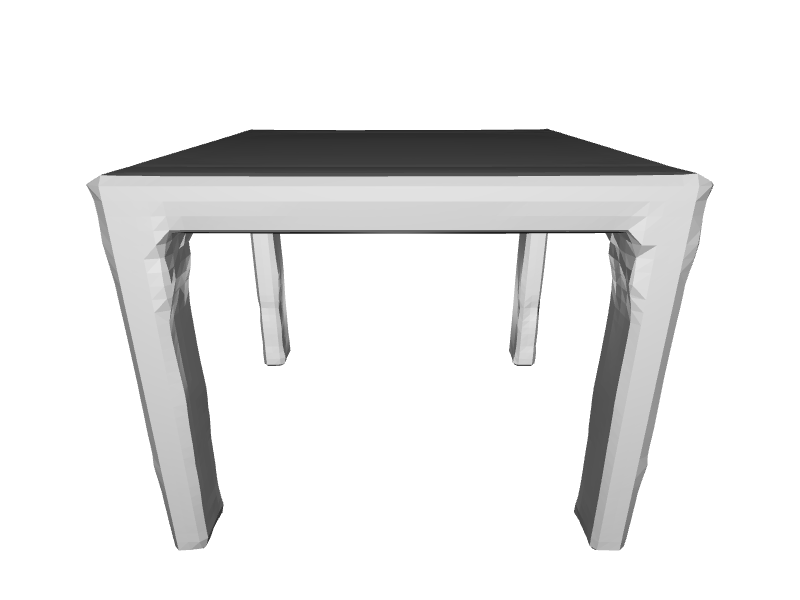}
    \includegraphics[scale=0.09,trim={2cm 1.5cm 2cm 1.5cm},clip]{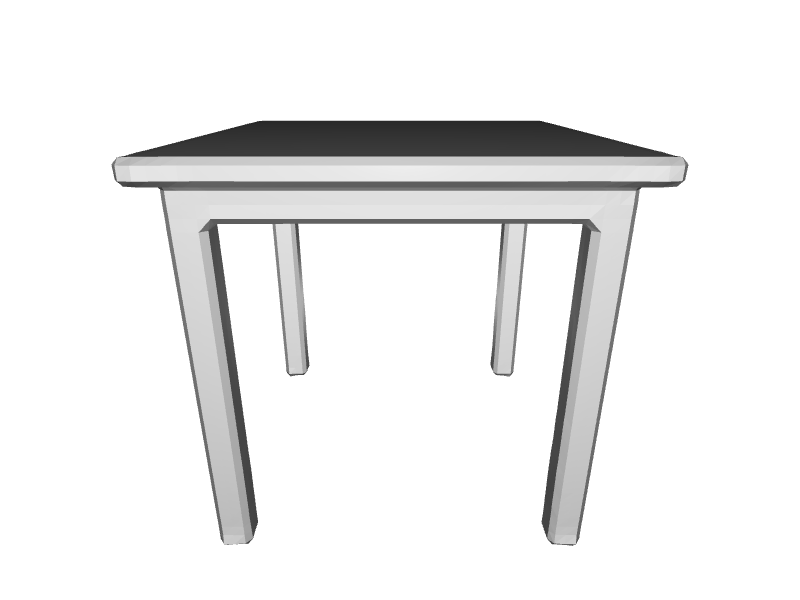}
	\includegraphics[scale=0.09,trim={2cm 1.5cm 2cm 1.5cm},clip]{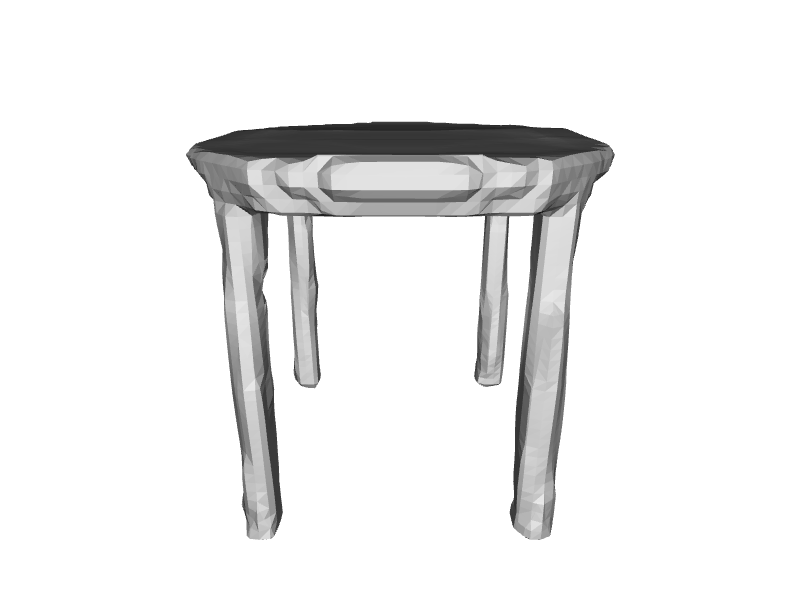}
	\includegraphics[scale=0.09,trim={2cm 1.5cm 2cm 1.5cm},clip]{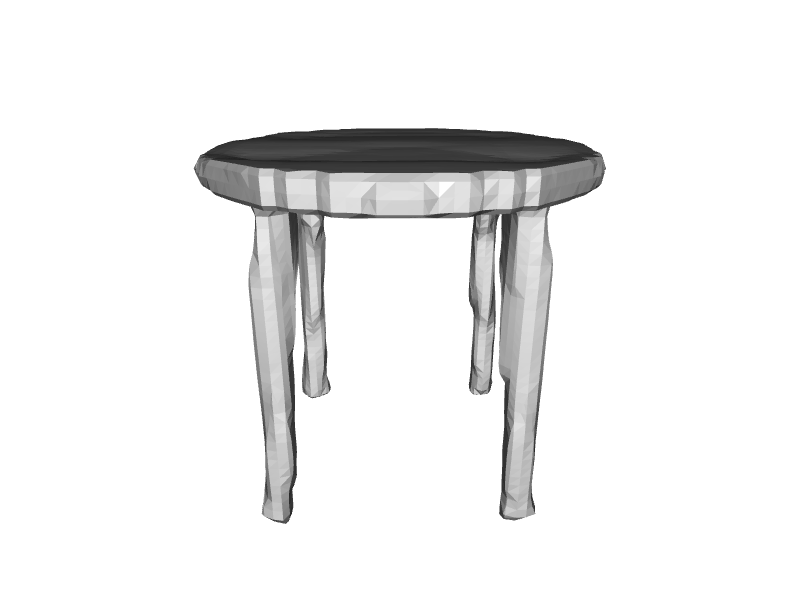} \\
	\rotatebox{90}{ \quad \ourH } &
    \includegraphics[scale=0.09,trim={2cm 1.5cm 2cm 1.5cm},clip]{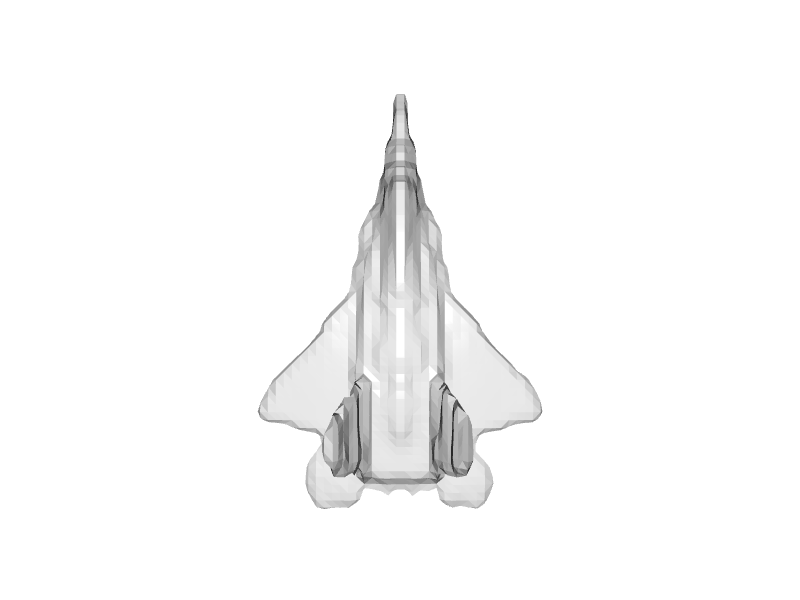}
    \includegraphics[scale=0.09,trim={2cm 1.5cm 2cm 1.5cm},clip]{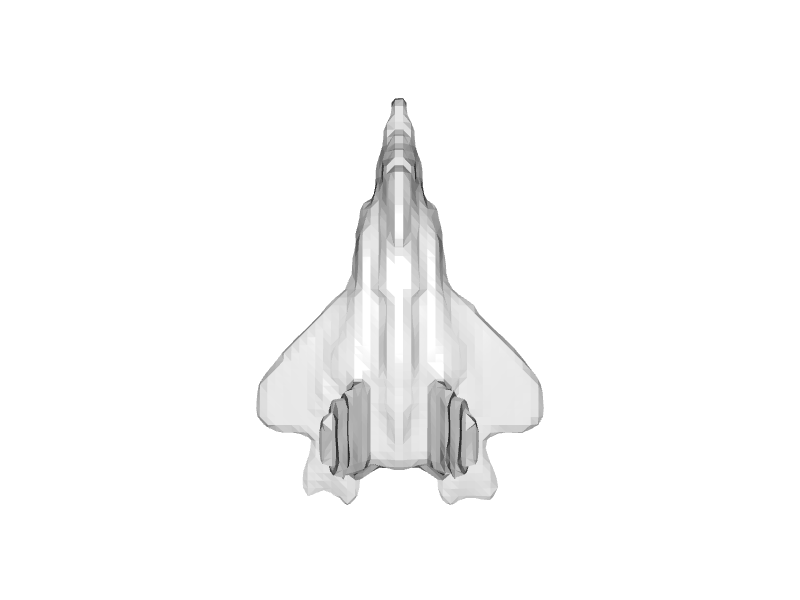}
	\includegraphics[scale=0.09,trim={2cm 1.5cm 2cm 1.5cm},clip]{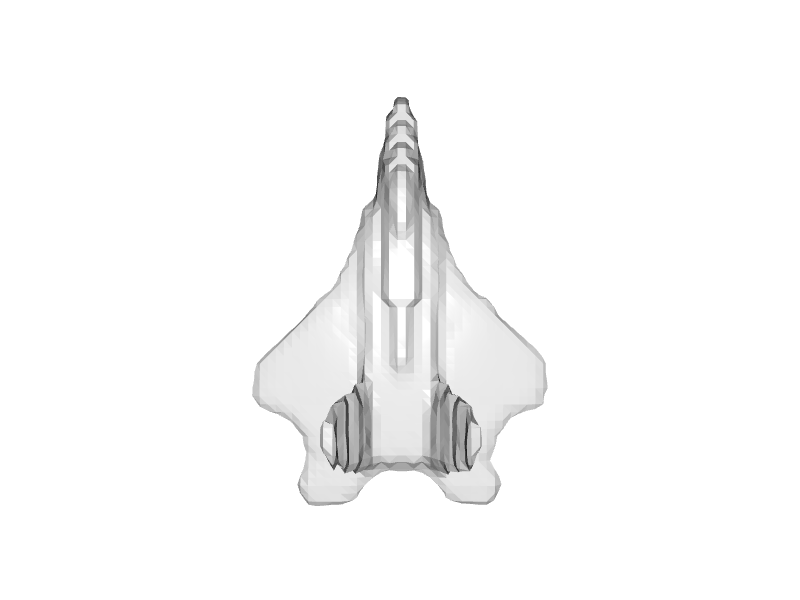}
	\includegraphics[scale=0.09,trim={2cm 1.5cm 2cm 1.5cm},clip]{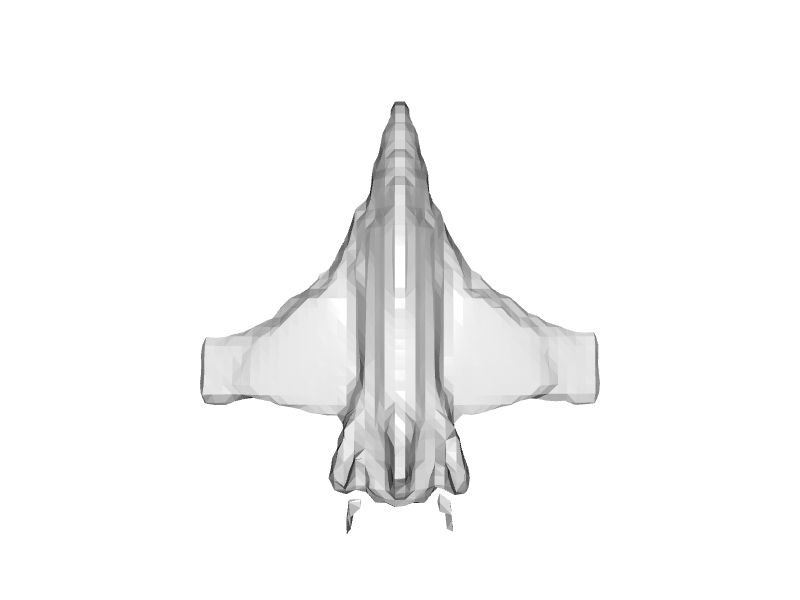}
	\includegraphics[scale=0.09,trim={2cm 1.5cm 2cm 1.5cm},clip]{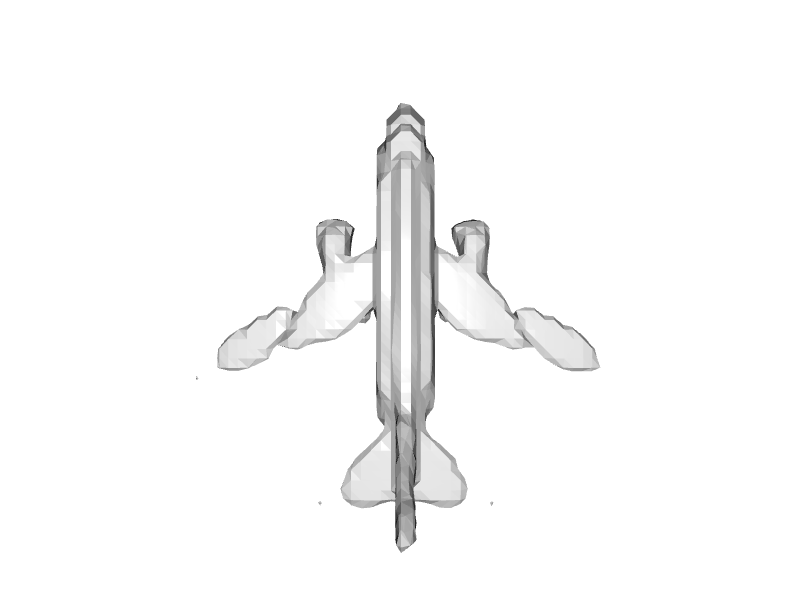} 
	\includegraphics[scale=0.09,trim={2cm 1.5cm 2cm 1.5cm},clip]{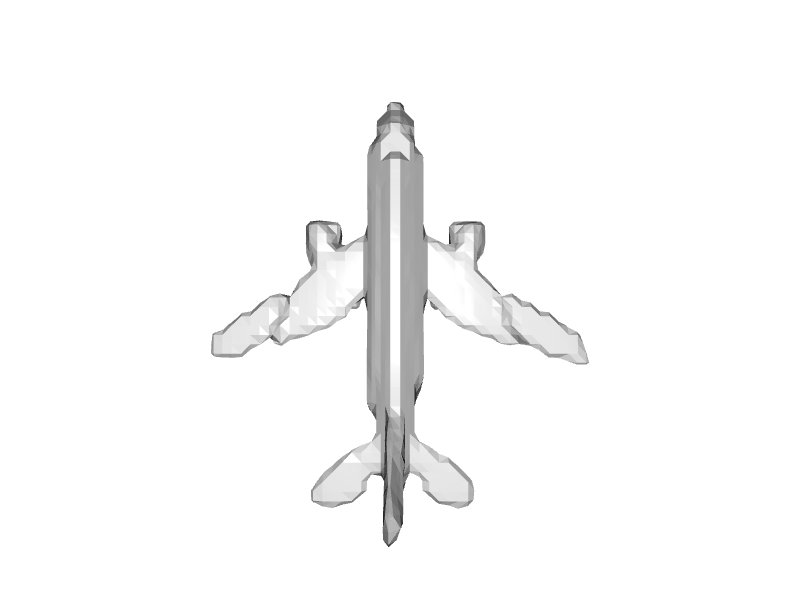} \\
	\rotatebox{90}{ \quad \ourH } &
	 \includegraphics[scale=0.09,trim={2cm 1.5cm 2cm 1.5cm},clip]{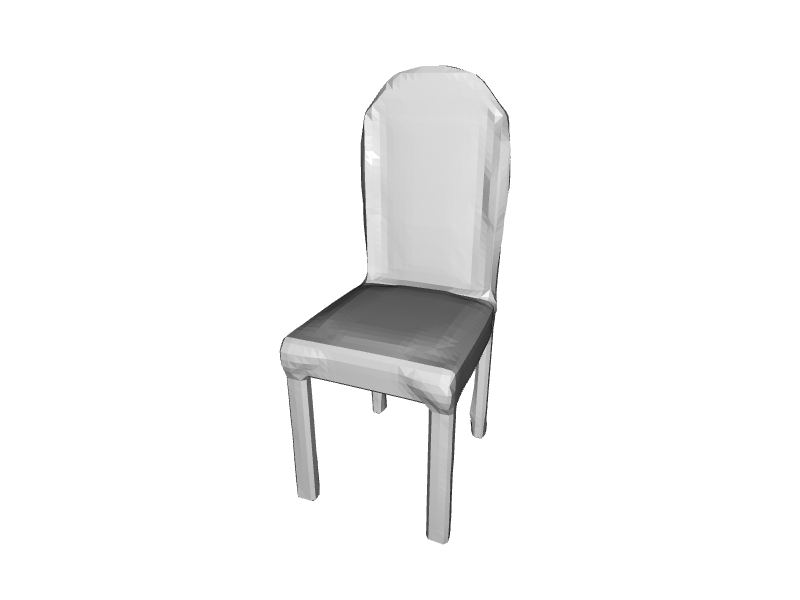}
    \includegraphics[scale=0.09,trim={2cm 1.5cm 2cm 1.5cm},clip]{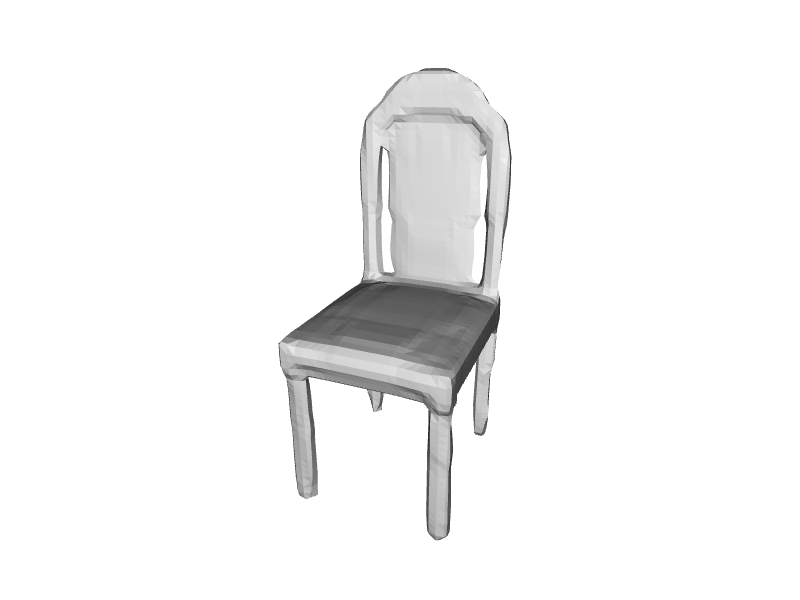}
    \includegraphics[scale=0.09,trim={2cm 1.5cm 2cm 1.5cm},clip]{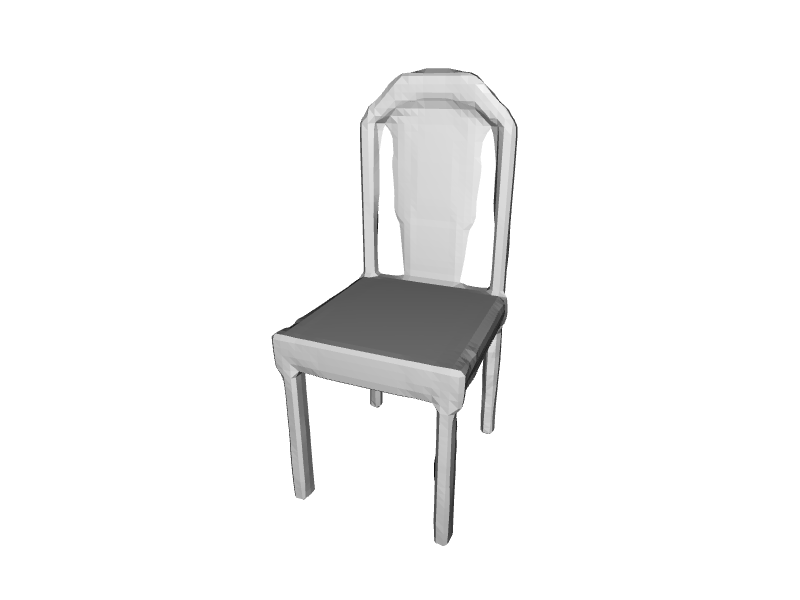}
	\includegraphics[scale=0.09,trim={2cm 1.5cm 2cm 1.5cm},clip]{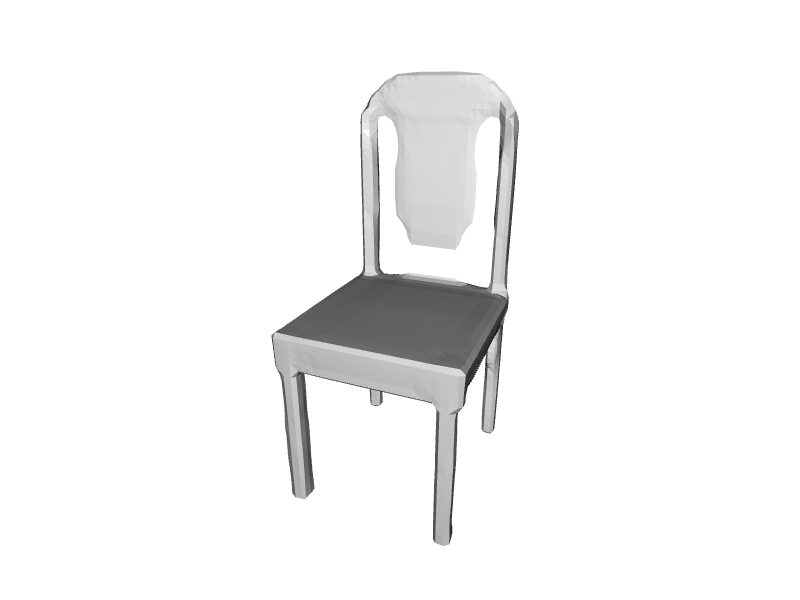}
	\includegraphics[scale=0.09,trim={2cm 1.5cm 2cm 1.5cm},clip]{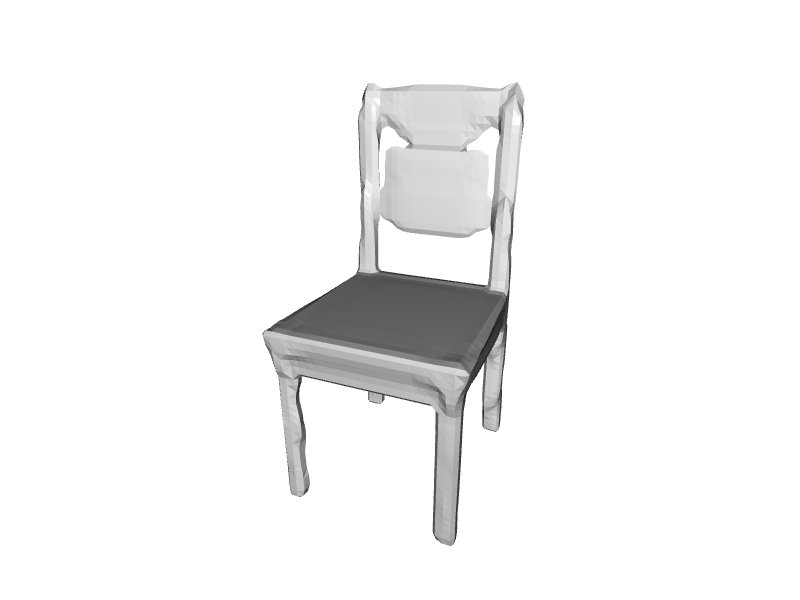}
	\includegraphics[scale=0.09,trim={2cm 1.5cm 2cm 1.5cm},clip]{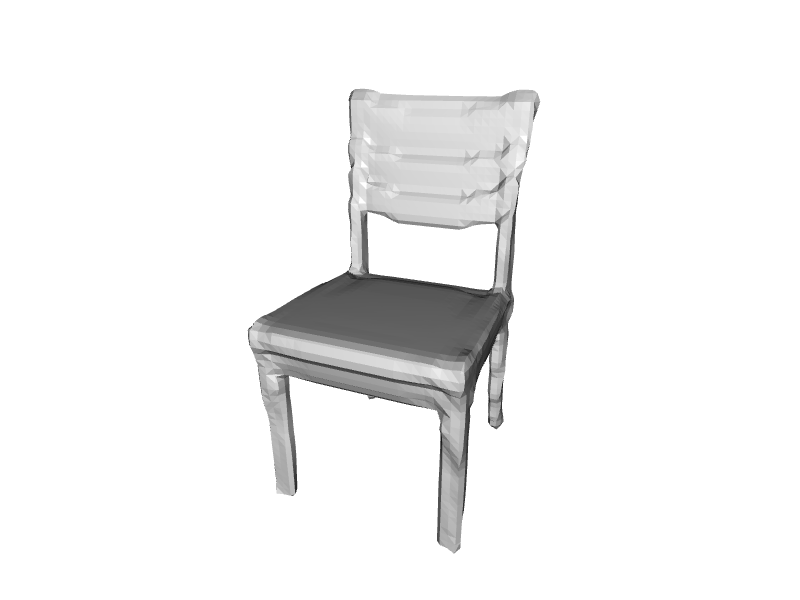} \\
	\rotatebox{90}{ \quad \ourH } &
	 \includegraphics[scale=0.09,trim={2cm 1.5cm 2cm 1.5cm},clip]{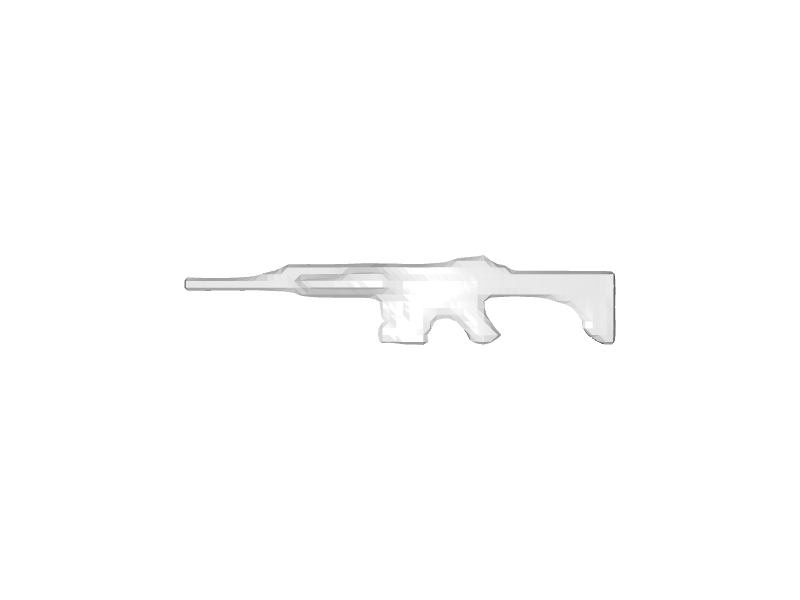}
    \includegraphics[scale=0.09,trim={2cm 1.5cm 2cm 1.5cm},clip]{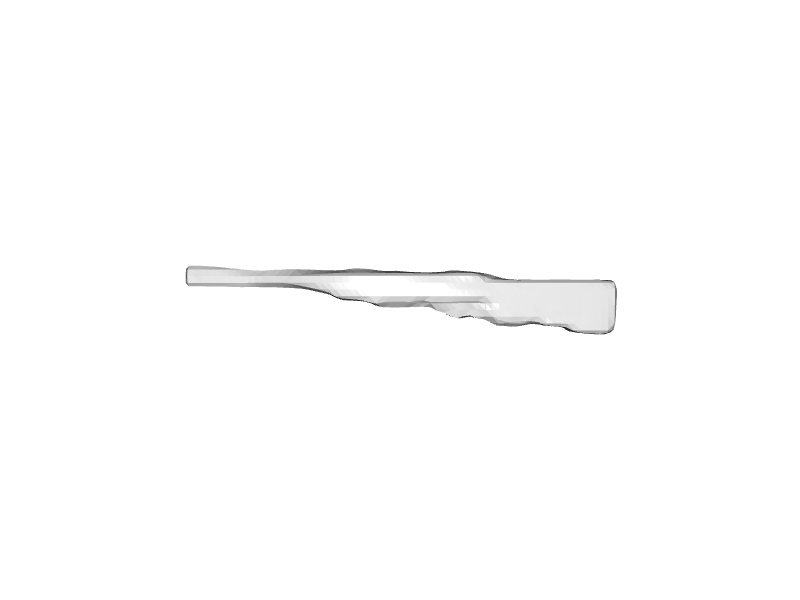}
    \includegraphics[scale=0.09,trim={2cm 1.5cm 2cm 1.5cm},clip]{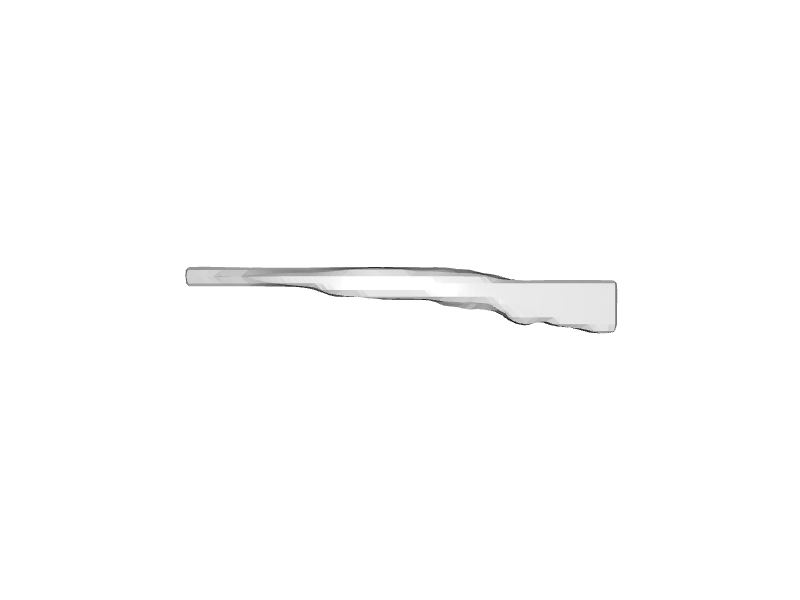}
	\includegraphics[scale=0.09,trim={2cm 1.5cm 2cm 1.5cm},clip]{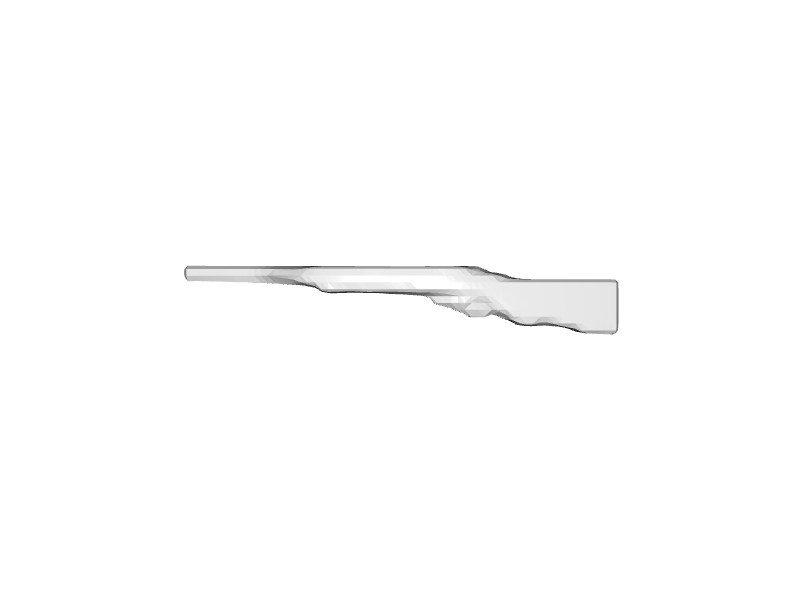}
	\includegraphics[scale=0.09,trim={2cm 1.5cm 2cm 1.5cm},clip]{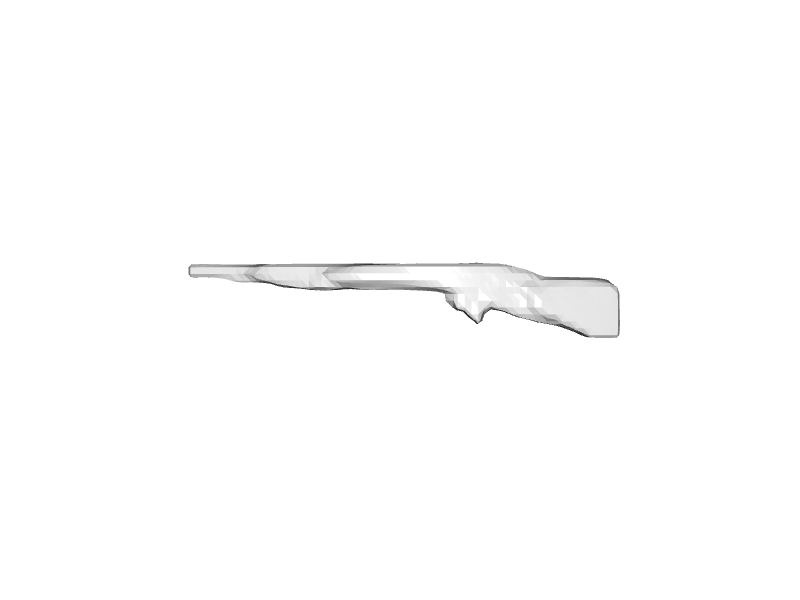}
	\includegraphics[scale=0.09,trim={2cm 1.5cm 2cm 1.5cm},clip]{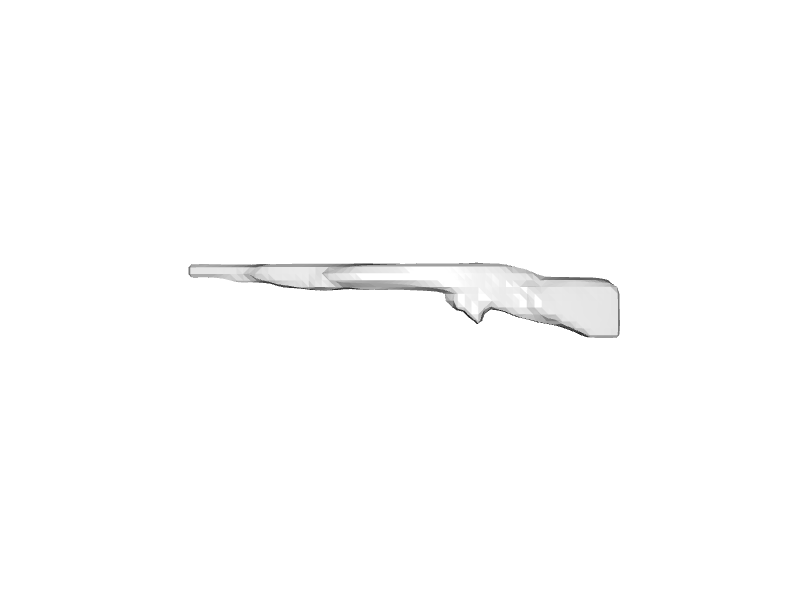}
	\end{tabular}
	\caption{Interpolations produce by \ourH{}. We take
two 3d objects and obtain a smooth transition between them.}
\label{fig:interpolations}
\end{figure}

\subsection{\ourH{}}

In this section, we introduce our~\ourH{} approach that draws the inspiration from the above methods to address the limitations of the baseline IM-NET model. 
We produce weights for a target network that describes a 3D object. Instead of transferring a prior distribution, as done in the HyperCloud, we transfer through a network a voxel grid to predict inside/outside label for each coordinate, see Fig.~\ref{fig:architectur_2}.

\ourH{} model uses a hyper network to output weights of a small neural network that creates a 3D voxel representation of a 3D object, instead of generating a directly inside/outside label to a fixed grid on a 3D cube. More specifically, we parametrize the surface of 3D objects as a function $f : \R^3 \to \{0, 1\}$, which returns ab inside/outside category, given a point from grid $(x, y, z)$. 
In other words, instead of producing a 3D voxel representation, we construct a model which create multiple neural networks (a different neural network for each object) that model the surfaces of the objects.

In practice, we have one large architecture (hyper network) that produces different weights for each 3D object (target network).
More precisely, we model function  $T_{\theta} : \R^3 \to \R$ (neural network classifier with weights $\theta$), which takes an element from the voxel grid and predicts an inside/outside label. 
In consequence, we can generate a 3D shape at any resolution by creating grids of different sizes and predict labels for its voxels.  

The target network is not trained directly.   
We use a hyper-network
$
\begin{array}{c}
H_{\phi}: \R^3 \supset X \to \theta ,
\end{array}
$
which for a voxel representation returns weights $\theta$ to the corresponding target network $T_{\theta}: X \to \{0,1\}$.
Thus, a 3D object is represented by a function (classifier) 
$$
\begin{array}{c}
T((x,y,x);\theta) = T((x,y,x); H_{\phi}(X)).
\end{array}
$$



More precisely, we take a voxel representation $X$ and pass it to $H_{\phi}$. The hyper network returns weights $\theta$ to target network $T_{\theta}$. Next, the input voxel representation $X$ (points sampled from $\X$) is compared with the output from the target network $T_{\theta}$ (we take voxel grid and predict inside/outside labels). 
To train our model we use mean squared error loss function.

We train a single neural model (hypernetwork), which allows us to produce a variety of functions at test time. By interpolating in the space of hypernetwork parameters, we are able to produce multiple shapes that bear similarity to each other, as Fig.~\ref{fig:interpolations} shows. 

The above architecture gives competitive qualitative and quantitative results to IM-NET, as we show in Section~\ref{sec:reco}, yet it offers a significant processing speedup confirmed by the results presented in Section~\ref{sec3:comple}. However, the remaining shortcoming of IM-NET, namely the reconstruction artifacts close to classification boundaries resulting from sampling strategy, remains. To address this limitation and process entire 3D cubes instead of sampled points, we leverage interval arithmetic~\cite{dahlquist2008numerical} and a neural architecture that implements it, IntervalNet~\cite{gowal2018effectiveness,morawiecki2019fast}. 

\subsection{Interval arithmetic}

Interval arithmetic allows to address the problems with precise numerical calculations caused by the rounding errors that appear within the computer representations of real numbers. In neural networks interval arithmetic is used to train models on interval data \cite{chakraverty2014interval,chakraverty2017novel,sahoo2020structural} and produce neural networks that are robust against adversarial attacks~\cite{gowal2018effectiveness,morawiecki2019fast}.

For the convenience of the reader, we give a short description of interval arithmetic. Interval arithmetic~\cite{dahlquist2008numerical}(Chapter 2.5.3) is based on the operations on segments. 
Let us assume $A$ and $B$ as numbers expressed as interval. For all $\bar a, \barbelow a, \bar b, \barbelow b \in R$ where 
$A = [\barbelow a, \bar a]$ , $B = [\barbelow b, \bar b]$, the main operations of intervals may be written as \cite{lee2004first}:
\begin{itemize}
    \item addition:
    $
     [\barbelow a, \bar a] +  [\barbelow b, \bar b] = [\barbelow a + \barbelow b, \bar a + \bar b]
    $
    \item subtraction:
    $
     [\barbelow a, \bar a] -  [\barbelow b, \bar b] = [\barbelow a - \barbelow b, \bar a - \bar b]
    $   
    \item multiplication:
    $
     [\barbelow a, \bar a] *  [\barbelow b, \bar b] = [\min( \barbelow a * \barbelow b, \barbelow a * \bar b, \bar a * \barbelow b, \bar a * \bar b),
     \max( \barbelow a * \barbelow b, \barbelow a * \bar b, \bar a * \barbelow b, \bar a * \bar b )
     ]
    $      
    \item division:
    $
     [\barbelow a, \bar a] \div [\barbelow b, \bar b] = [\min( \barbelow a \div \barbelow b, \barbelow a \div \bar b, \bar a \div \barbelow b, \bar a \div \bar b),
     \max( \barbelow a \div \barbelow b, \barbelow a \div \bar b, \bar a \div \barbelow b, \bar a \div \bar b )
     ]
    $
    excluding the case $\barbelow b = 0$ or $ \bar b = 0$.

\end{itemize}

The above operations allow the construction of a neural network, which uses intervals instead of points on neural networks input. The following subsection shows that we can substitute classical arithmetic on points to the arithmetic on intervals and, consequently, directly work on voxels.

\begin{figure}[h!]
	\centering
	\includegraphics[scale=0.35]{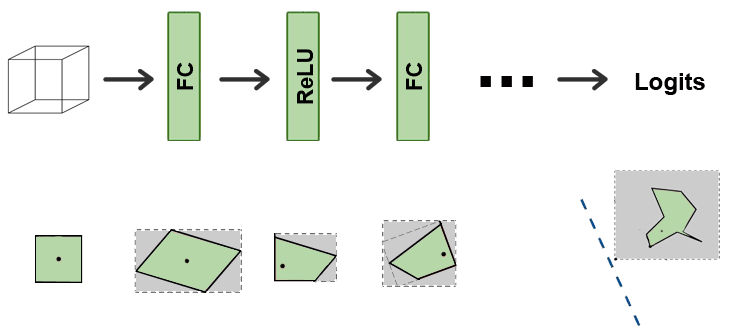}
	\caption{ Illustration of IntervalNet 3D interval (illustrated in 2D for clarity) is propagated through MLP. At each layer, the 3D cube is deformed by linear transformation (FC layer). Then we construct interval bound (marked by gray color) and use the ReLU activation function. Interval bound (in gray) is propagated to the next layer. In logit space, it becomes easy to compute an upper bound on the worstcase violation. Interval bound (in gray) is moved on one side of the decision boundary and consequently transformed 3D cube (in green) is correctly classified. The figure is inspired by Fig.~2 from \cite{gowal2018effectiveness}.  }
\label{fig:intervalnet}
\end{figure}


\subsection{\our{}}

Thanks to the interval arithmetic, we can construct IntervalNet~\cite{gowal2018effectiveness,morawiecki2019fast} that is able to process 3D cubes defined as intervals in the 3D space.
Let us consider the feed-forward neural network trained for a classification task. 
We assume that the network is defined by a sequence of transformations $h_k$ for
each of its $K$ layers. That is, for input $z_0$, we have
$$
z_k = h_k(z_{k - 1}) \mbox{ for } k = 1, \ldots, K. 
$$
The output $z_K \in \R^N$ has $N$ logits corresponding to $N$
classes.
In the IntervalNet each transformation $h_k$ uses interval arithmetic. More precisely, the input of a layer is a 3D interval (3D cube).

Let us consider a classical dense layer:
$$
h(x) = W x + b.
$$
In IntervalNet, the input of the network is an interval and consequently the output of a dense layer is also a vector of intervals:
$$
[ \barbelow h(x), \bar h(x)] = W [ \barbelow x, \bar x ] + b.
$$
Therefore, the IntervalNet is defined by a sequence of transformations $h_k$ for
each of its $K$ layers. That is, for an input $[\barbelow x, \bar x ]$, we have
$$
[\barbelow z_k, \bar z_k] = h_k([\barbelow z_{k-1}, \bar z_{k-1}]) \mbox{ for } k = 1, \ldots, K. 
$$
The output $[\barbelow z_K, \bar z_K] $ has $N$ interval logits corresponding to $N$
classes.
Propagating intervals through any element-wise monotonic activation function (e.g., ReLU, tanh, sigmoid) is straightforward. If $h_k$ is an element-wise increasing function, we
have:
$$
\barbelow z_{k} = h_{k}( \barbelow z_{k-1}),  \qquad
\bar z_{k} = h_{k}( \bar z_{k-1}).
$$

\begin{figure}
    \centering
    \includegraphics[width=0.49\linewidth]{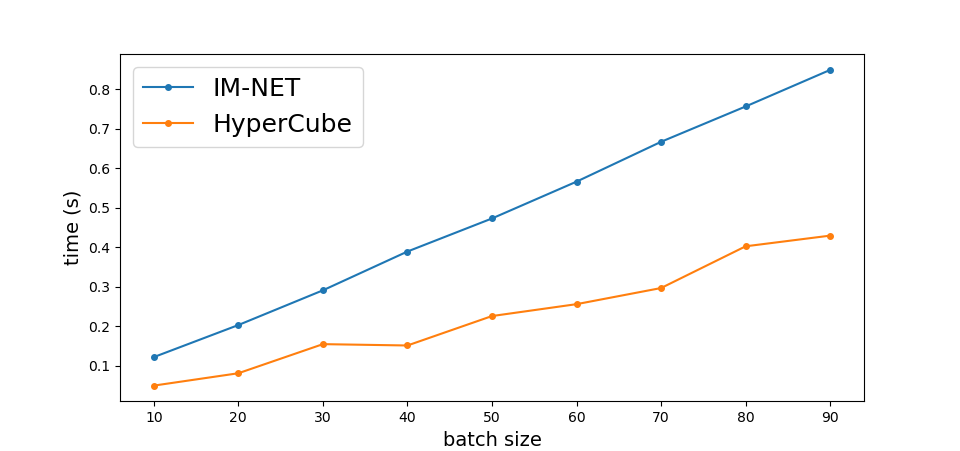}
    \includegraphics[width=0.49\linewidth]{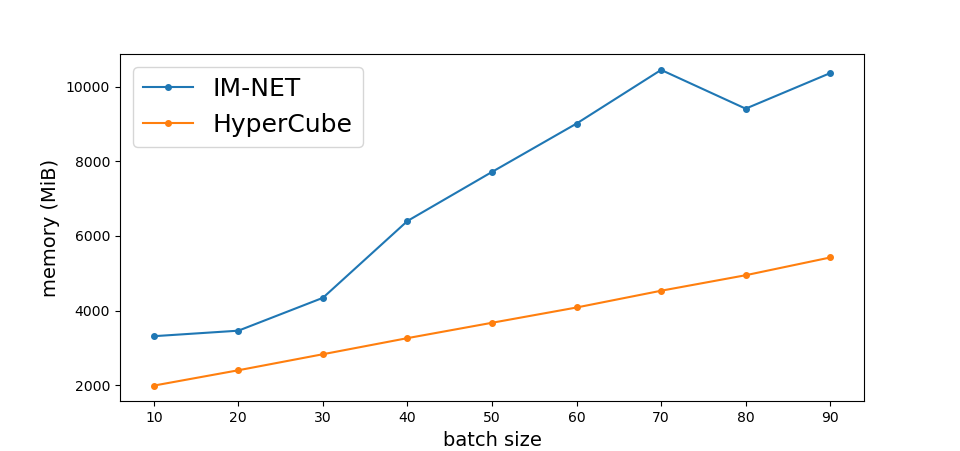}    
    \caption{Comparison of training times and GPU memory used by IM-NET and \ourH{}. Our \ourH{} method offers over an order of magnitude decrease in both training time and memory.}
    \label{fig:time}
\end{figure}

Our goal is to obtain a classification rate for all elements from the interval. Therefore, we consider the worst-case prediction for the whole interval bound $[\barbelow z_K, \bar z_K] $ of the final logits. More precisely, we need to ensure that the entire bounding box is classified correctly, i.e., no perturbation changes the correct class label. In consequence, the logit of the true class is equal to its lower bound, and the other logits are equal to their upper bounds:
$$
\hat{z}_{K,y} = \left\{
\begin{array}{ll}
     \overline{z}_{K,y}, & \mbox{ for } y \neq y_{true},\\[0.8ex]
     \underline{z}_{K,y_{true}}, & \mbox{ otherwise }.
\end{array}
\right.
$$
Finally, one can apply softmax with the cross-entropy loss to the logit vectors $\hat{z}_K$ representing the worst-case prediction, see Fig.~\ref{fig:intervalnet}. For more detailed information see~\cite{gowal2018effectiveness,morawiecki2019fast}.

Such architecture can be simply added to \ourH{} as a target network.
It turns out that the output of the interval layer can be calculated only by two matrix multiplies:
$$
\begin{array}{lll}
    & z_{k-1} = \frac{\bar z_{k-1} + \barbelow z_{k-1}}{2},
    & r_{k-1} = \frac{\bar z_{k-1} - \barbelow z_{k-1}}{2},\\[0.8ex]
    & z_k = W_k z_{k-1} + b_{k-1},
    & r_k = |W_k| r_{k-1},\\[0.8ex]
    & \barbelow z_k = z_k - r_k,
    & \bar z_k = z_k + r_k,\\
\end{array}
$$
where $| \cdot |$ is the element-wise absolute value operator and $r_{k}$ is a radius of the interval.

Consequently, IntervalNet has exactly the same number of weights as in a fully connected architecture with the some architecture. \our{} is a copy of \ourH{} with using IntervalNet instead of MLP as a target network. We use worst-case accuracy instead of mse to ensure that the whole voxel is correctly classified.

It should be highlighted that IntervalNet can not be added to IM-NET since we concatenate feature vectors with 3D coordinates. More precisely, feature vectors must be propagated by fully connected architecture and 3D cube by interval arithmetic, which is hard to implement in one neural network.

\section{Experiments}

In this section, we describe the experimental results of the
proposed generative models in various tasks, including 3D
mesh generation and interpolation. We show that our model is essentially faster and requires less memory in training time. Then, we compare our model with a baseline on reconstruction and generative tasks.

\begin{table}[h!]
    \centering
    \begin{tabular}{ccccccc}
         \toprule
                                                  &         & Plane & Car & Chair & Rifle & Table \\
         \midrule
         \multicolumn{1}{c}{\multirow{2}{*}{MSE}} & IM-NET  & \textbf{2.14} & 4.99 & 11.43 & \textbf{1.91} & 10.67 \\
                                                  & \ourH{} & 2.44 & \textbf{4.37} & \textbf{9.07} & \textbf{1.91} & \textbf{9.37} \\
         \midrule
         \multicolumn{1}{c}{\multirow{2}{*}{IoU}} & IM-NET  & \textbf{78.77} & 89.26 & 65.65 & \textbf{72.88} & 71.44 \\
                                                  & \ourH{} & 65.35 & \textbf{90.05} & \textbf{72.61} & 63.97 & \textbf{73.78} \\
         \midrule
         \multicolumn{1}{c}{\multirow{2}{*}{CD}}  & IM-NET &  \textbf{4.23} & 5.44 & 9.05 & \textbf{3.77} & 11.54 \\ 
                                                  & \ourH{} & 4.74 & \textbf{3.36} & \textbf{8.35} & 4.20 & \textbf{8.82} \\
         \bottomrule
    \end{tabular}
    \caption{The mean is calculated for reconstructions of 100 first elements from test set in each category. MSE is multiplied by $10^3$, IoU by $10^2$, and CD by $10^4$. }
    \label{tab:my_label}
\end{table}




\subsection{Training time and memory footprint comparison}\label{sec3:comple}

Fig.~\ref{fig:time} displays a comparison between our \ourH{} method and the competing IM-NET. For a fair comparison we evaluated the architectures proposed in~\cite{chen2019learning} . The models were trained on the ShapeNet dataset. Our \ourH{} approach leads to a significant reduction in both training time and memory footprint due to a more compact architecture.

\subsection{Reconstruction capabilities }
\label{sec:reco}

For the quantitative comparison of our method with the current state-of-the-art solutions in the reconstruction task, we follow the approach introduced in~\cite{chen2019learning}. Metrics for encoding and reconstruction are based on point-wise distances, e.g., Chamfer Distance (CD), Mean Squared Error (MSE) and Intersection over Union (IoU) on voxels.


The results are presented in Table~\ref{tab:my_label}. \ourH{} obtain comparable results to the reference method.

\begin{table}[h!]
    \centering
    \begin{tabular}{ccccccc}
        \toprule
         &         & Plane & Car & Chair & Rifle & Table \\
        \midrule
         \multicolumn{1}{c}{\multirow{4}{*}{MSE}} & IM-NET & \textbf{2.98} &10.98 & 17.11 & \textbf{2.41}  & 13.38 \\
         & IM-NET \% & 0.45 & \textbf{0.70} & 0.68 & 0.56 & 0.64 \\
         & \ourH{} & 2.99 & \textbf{7.47} & \textbf{16.46} & 2.61 & \textbf{13.23} \\
         & \ourH{} \% &  \textbf{0.57} & \textbf{0.70} & \textbf{0.72} & \textbf{0.68} &  \textbf{0.69} \\
         \midrule
         \multicolumn{1}{c}{\multirow{4}{*}{IoU}} & IM-NET & 56.05 & 77.36 & 50.46 &  51.53 &  54.08 \\
         & IM-NET \% & 0.61 & \textbf{0.71} &  0.72 & 0.69 &  0.75 \\
         & \ourH{} & \textbf{61.68} & \textbf{86.34} & \textbf{53.52} & \textbf{59.80} & \textbf{61.23} \\
         & \ourH{} \% & \textbf{0.67} & \textbf{0.71} & \textbf{0.76} & \textbf{0.73} & \textbf{0.77} \\
         \midrule
         \multicolumn{1}{c}{\multirow{4}{*}{CD}} & IM-NET & 7.38 & 5.72 & 13.99 & 8.06 &  17.36 \\
         & IM-NET & 0.58 & 0.71 & \textbf{0.78} & 0.72 &  \textbf{0.82}\textbf{} \\
         & \ourH{}  & \textbf{5.02} & \textbf{4.28} & \textbf{12.92} & \textbf{4.96} &  \textbf{12.49} \\
         & \ourH{} \% & \textbf{0.66} & \textbf{0.74} & \textbf{0.78} & \textbf{0.74} & 0.81 \\
         \bottomrule
    \end{tabular}
    \caption{Generation. Mean of minimum MSE/CD, maximum IoU between test set and 5*test\_set\_size generated objects, \% of test set objects matched as the closest ones.}
    \label{tab:my_label}
\end{table}
\subsection{Generative model}

We examine the generative capabilities of the provided
\ourH{} model compared to the existing reference IM-NET. 
Our model similarly to IM-NET can be used for generating 3D objects. For 3D shape generation, we employed latent-GANs~\cite{dong2014learning} on feature vectors learned by a 3D autoencoder. By analogy to IM-NET we did
not apply traditional GANs trained on voxel grids since the
training set is considerably smaller compared to the size of
the output. Therefore, the pre-trained AE serves as a dimensionality reduction method, and the latent-GAN
was trained on high-level features of the original shapes. 
In Table~\ref{tab:my_label} we present compression between IM-NET and \ourH{}. As we can see we obtain similar results than classical approaches.

\begin{figure}[h!]
	\centering
	\begin{tabular}{@{}cc@{}c@{}c@{}c@{}}
	\rotatebox{90}{ \quad input} &
	\includegraphics[scale=0.09,trim={2cm 0cm 2cm 0cm},clip]{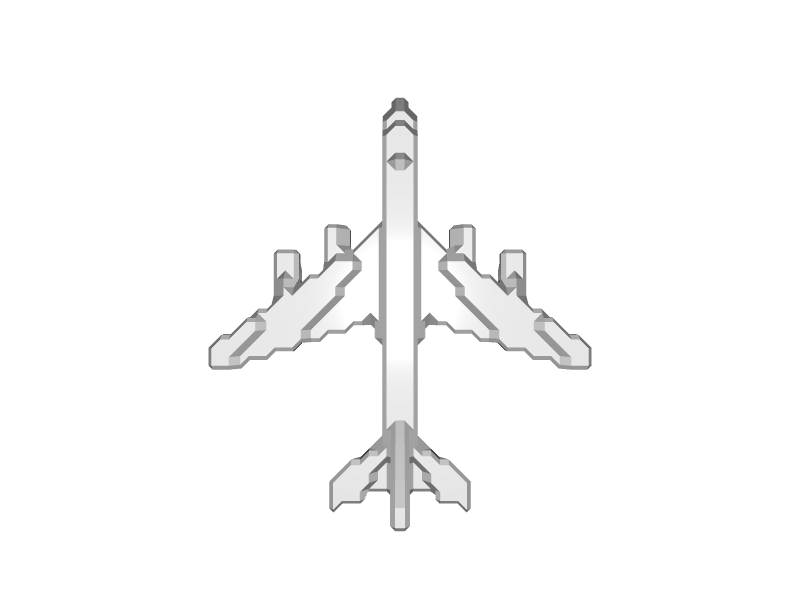}
    \includegraphics[scale=0.09,trim={2cm 0cm 2cm 0cm},clip]{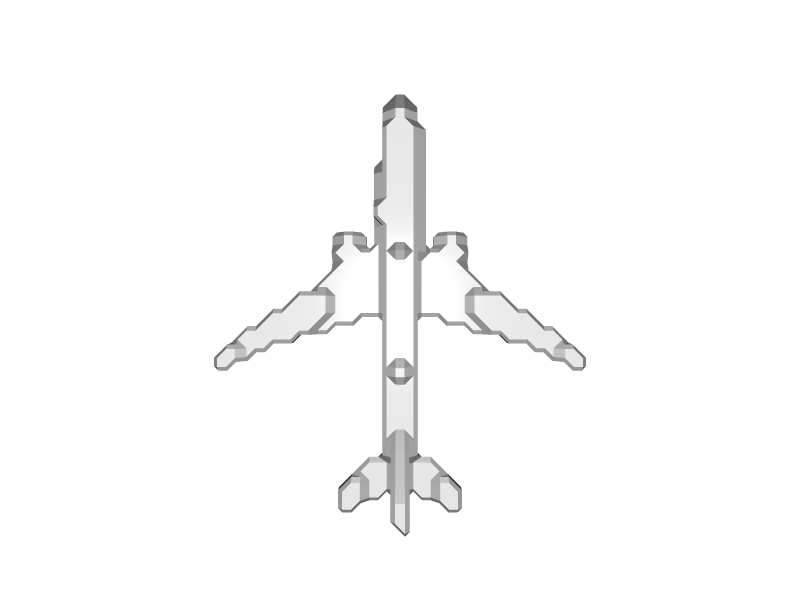}
    \includegraphics[scale=0.09,trim={2cm 0cm 2cm 0cm},clip]{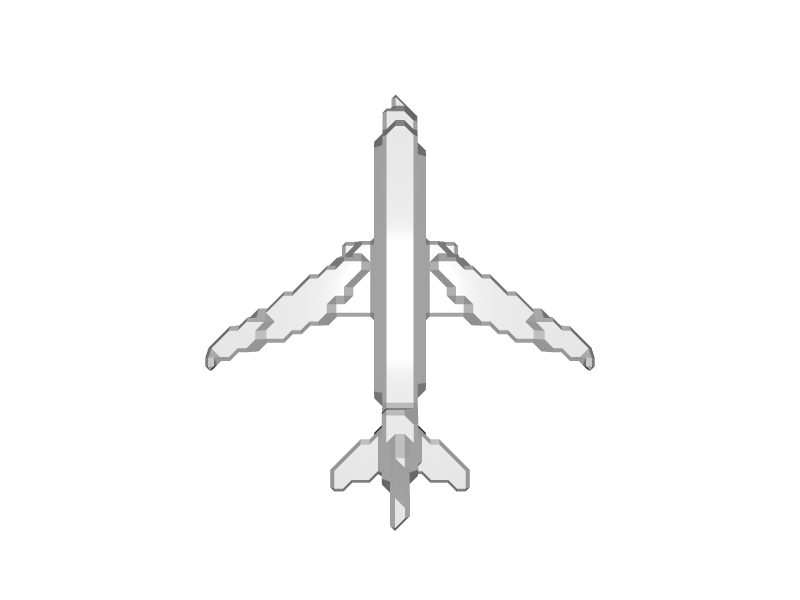}
    \includegraphics[scale=0.09,trim={2cm 0cm 2cm 0cm},clip]{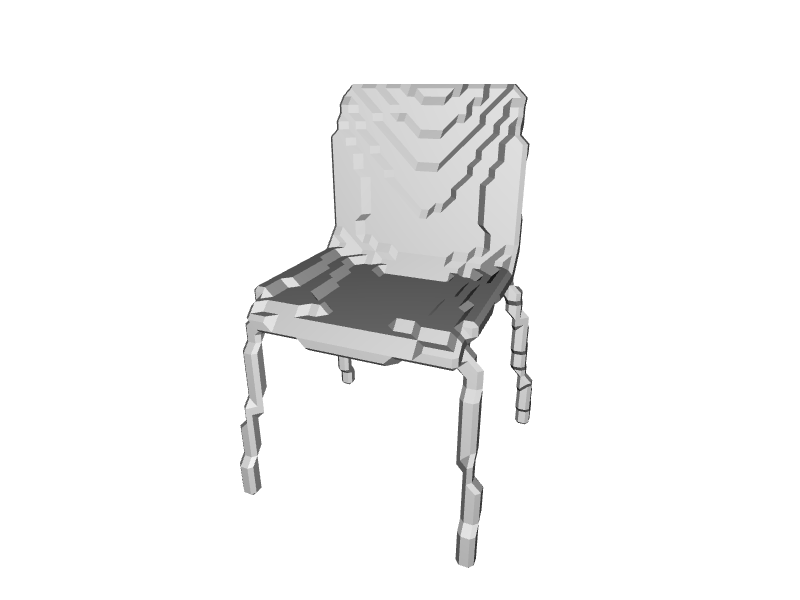}
    \includegraphics[scale=0.09,trim={2cm 0cm 2cm 0cm},clip]{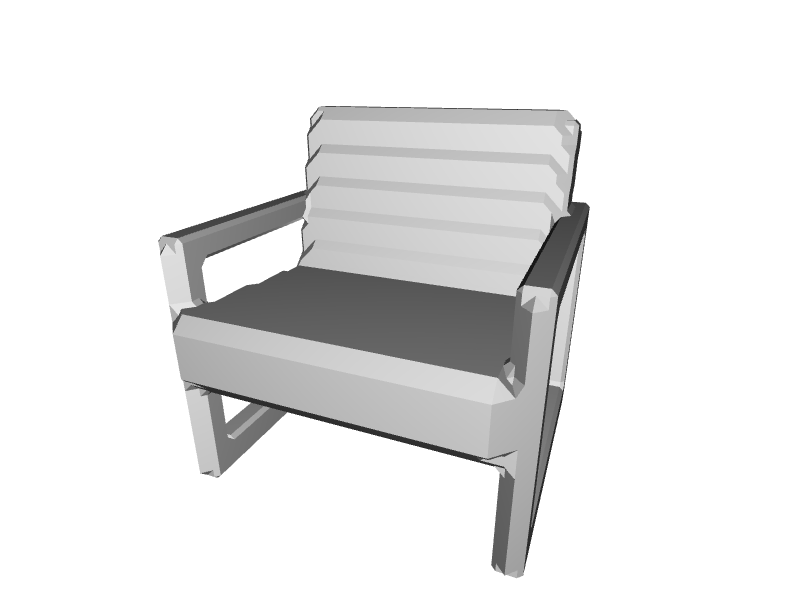}
    \includegraphics[scale=0.09,trim={2cm 0cm 2cm 0cm},clip]{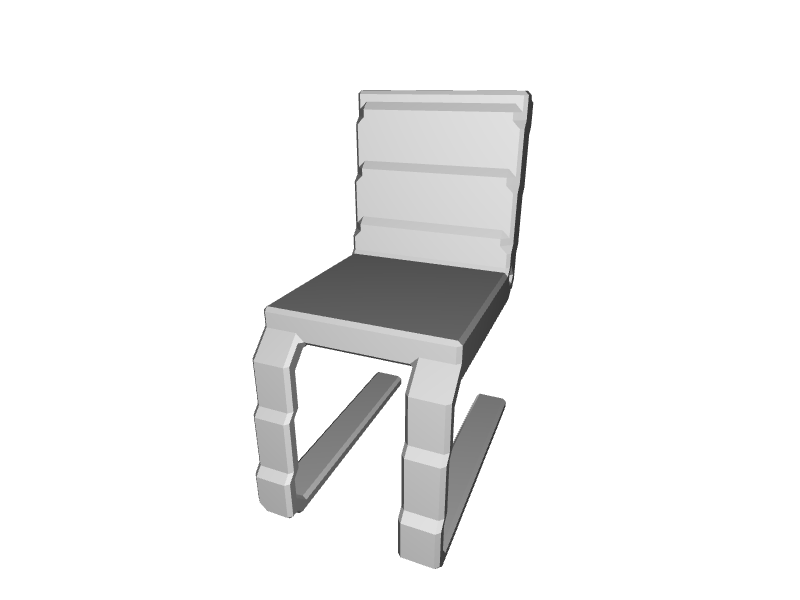}
    \\
    \rotatebox{90}{ \quad \ourH{} }  &
	\includegraphics[scale=0.09,trim={2cm 0cm 2cm 0cm},clip]{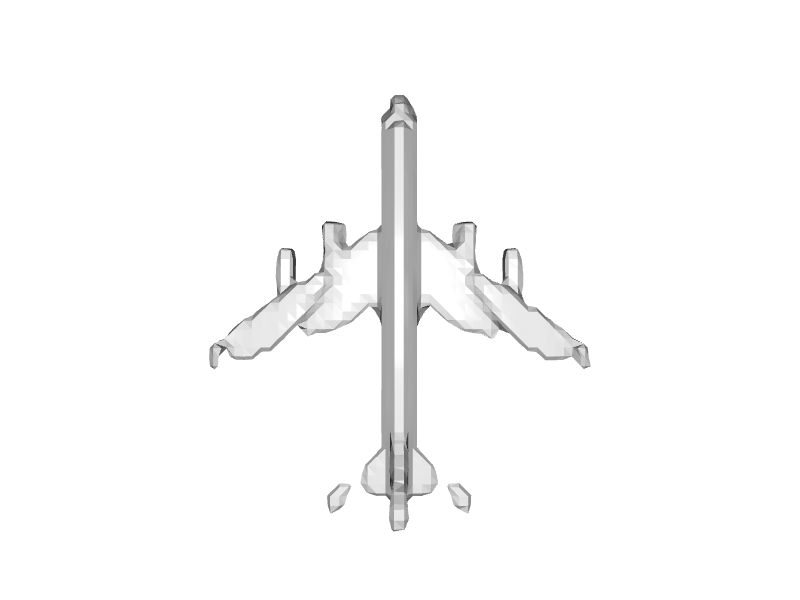}
    \includegraphics[scale=0.09,trim={2cm 0cm 2cm 0cm},clip]{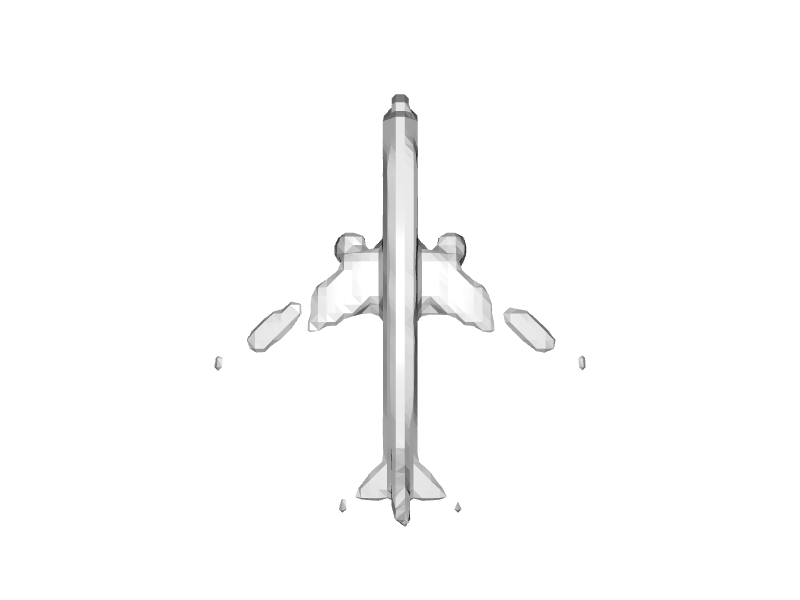}
    \includegraphics[scale=0.09,trim={2cm 0cm 2cm 0cm},clip]{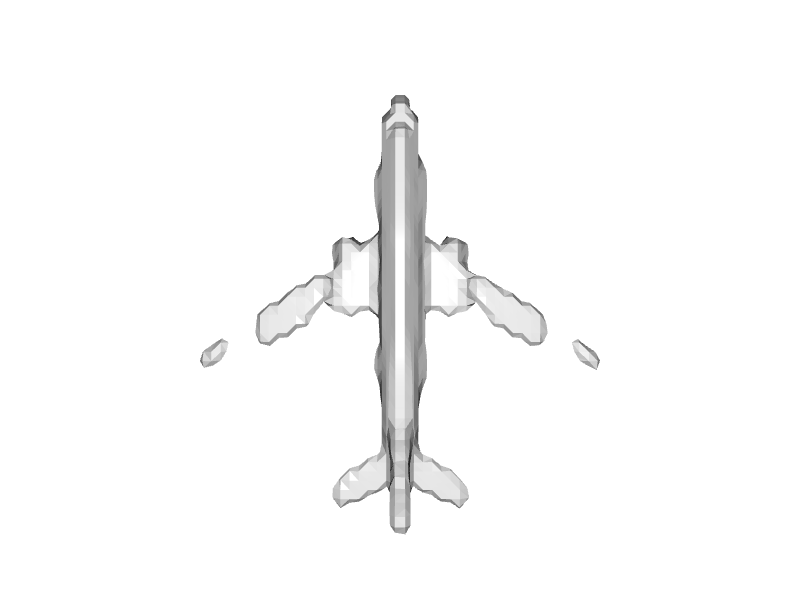}
    \includegraphics[scale=0.09,trim={2cm 0cm 2cm 0cm},clip]{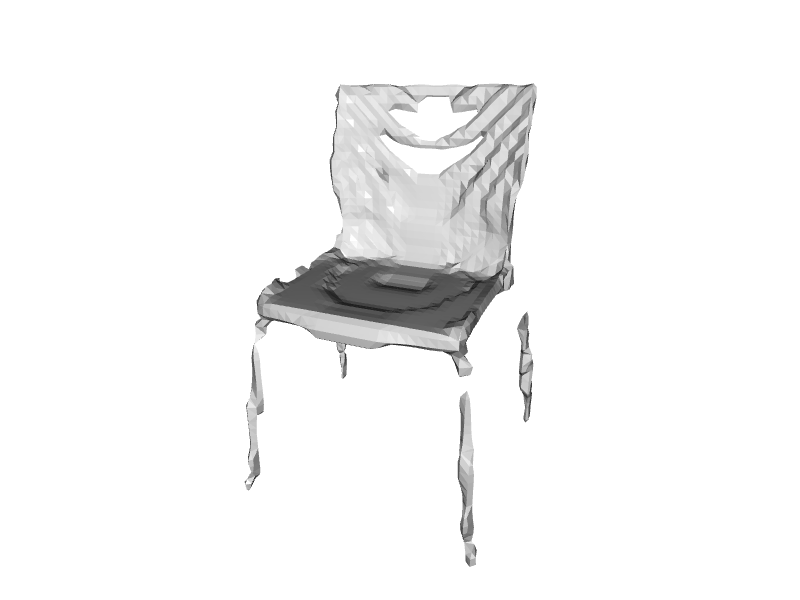}
    \includegraphics[scale=0.09,trim={2cm 0cm 2cm 0cm},clip]{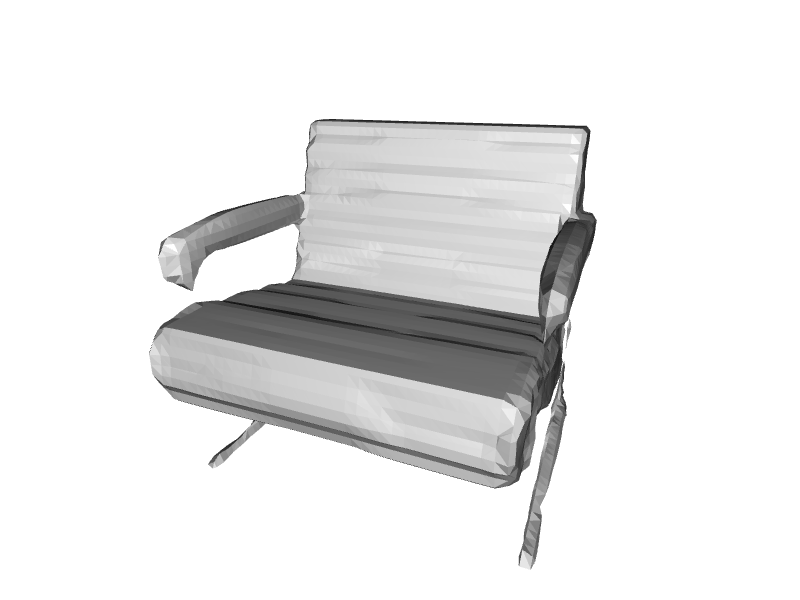}
    \includegraphics[scale=0.09,trim={2cm 0cm 2cm 0cm},clip]{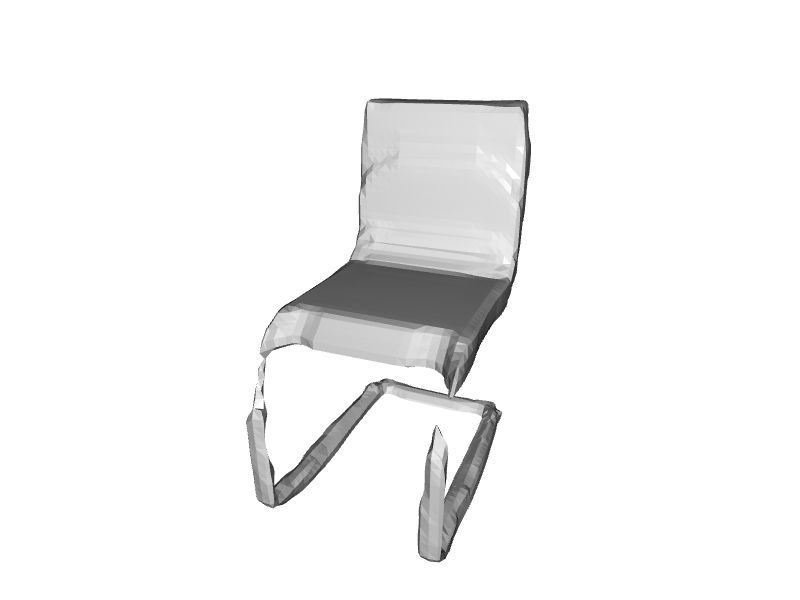}
    \\
    \rotatebox{90}{ \quad  \ourmultiline } &
	\includegraphics[scale=0.09,trim={2cm 0cm 2cm 0cm},clip]{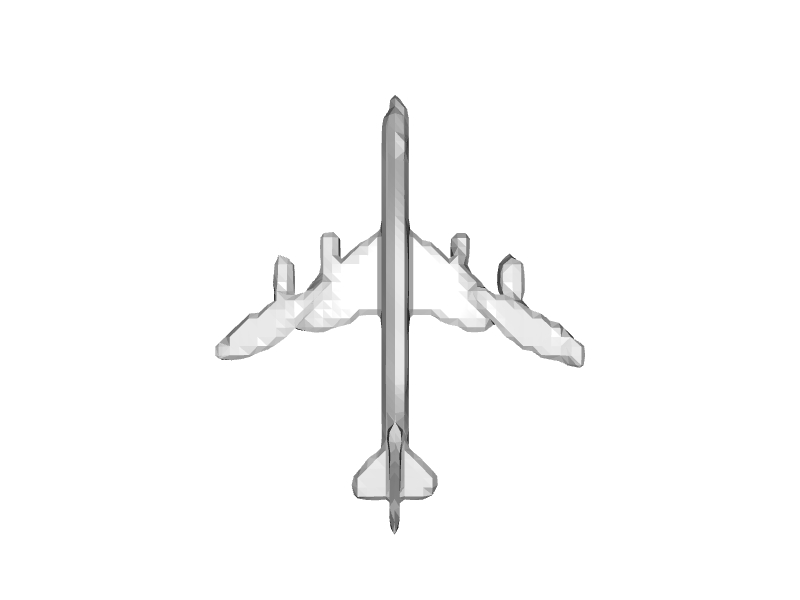}
    \includegraphics[scale=0.09,trim={2cm 0cm 2cm 0cm},clip]{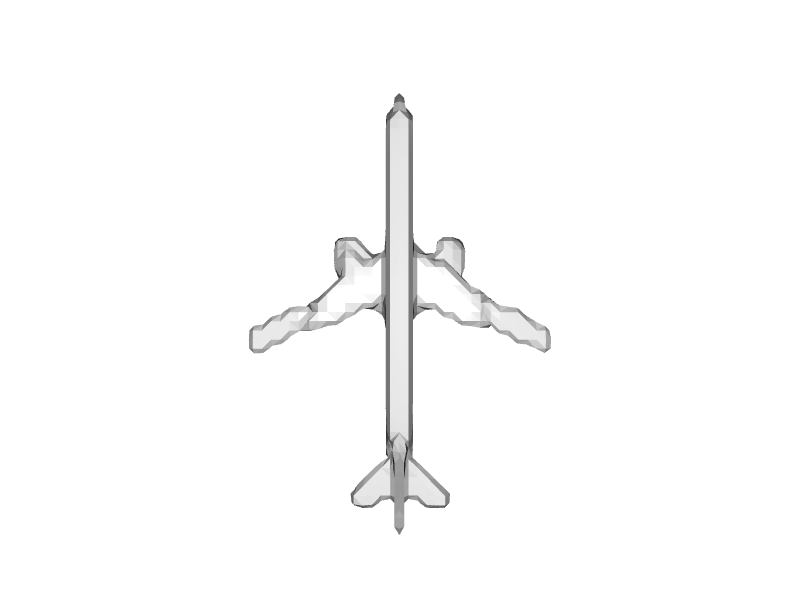}
    \includegraphics[scale=0.09,trim={2cm 0cm 2cm 0cm},clip]{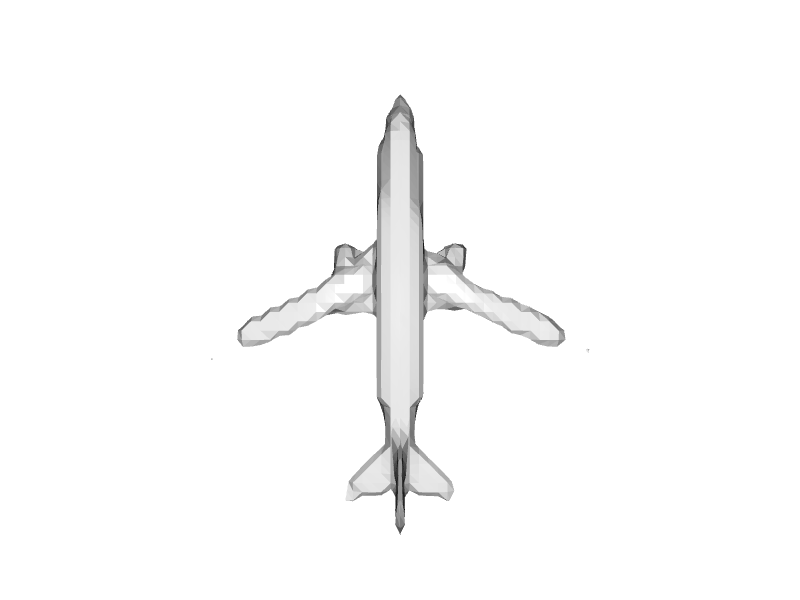}
    \includegraphics[scale=0.09,trim={2cm 0cm 2cm 0cm},clip]{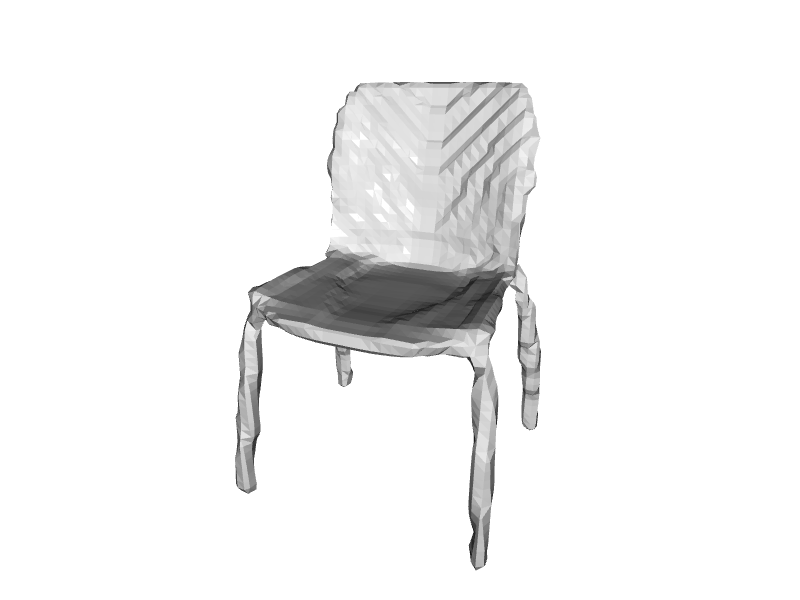}
    \includegraphics[scale=0.09,trim={2cm 0cm 2cm 0cm},clip]{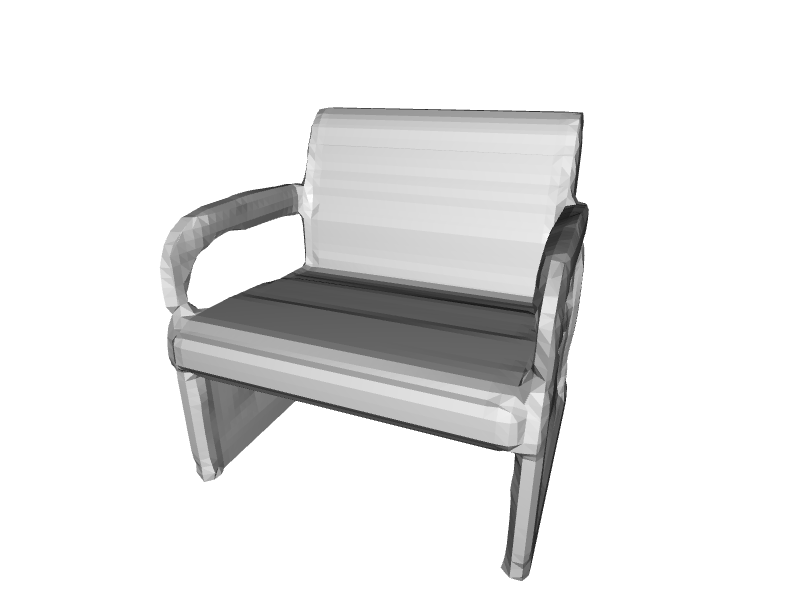}
    \includegraphics[scale=0.09,trim={2cm 0cm 2cm 0cm},clip]{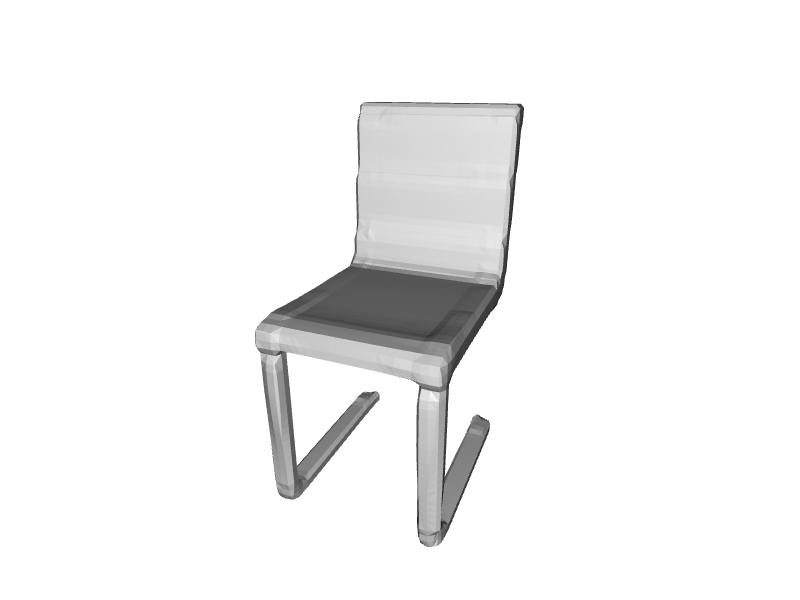}
		\end{tabular}
	\caption{Competition between models between architectures working on points (\ourH{}) and interval architecture (\our{}). As we can see, IntervalNet can fill some empty spaces in meshes.    }
	\label{fig:missing}
\end{figure}

\begin{figure}[t!]
	\centering
    \includegraphics[scale=0.25]{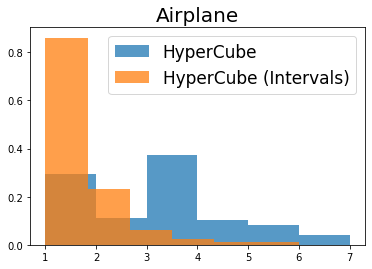}
    \includegraphics[scale=0.25]{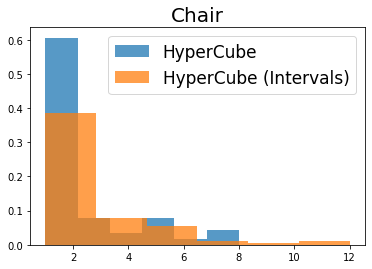}
    \includegraphics[scale=0.25]{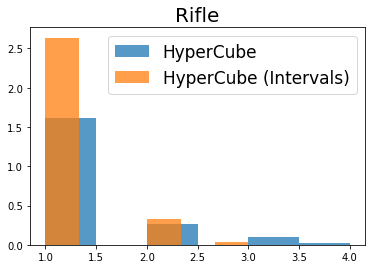}
    \includegraphics[scale=0.25]{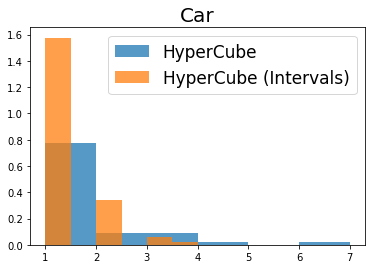}    
	\caption{ We calculate the number of connected components produced by mesh obtained by architecture with intervals and without.  Our \our{} approach provides better models for classes Airplane, Rifle, Car. }
	\label{fig:hist}
\end{figure}

\subsection{Intervals vs Points}

As it was shown in previous section, \our{} is essentially faster than IM-NET. Both architectures give similar reconstruction and generative capability. Since \ourH{} as well as IM-NET works directly on points the classification boundary is not smooth and we can see artifacts in mesh reconstruction, see Fig.~\ref{fig:missing}. Since mesh representation is produced by the Marching Cubes algorithm such artefacts appear with object with similar value of MSE, CD, and IoU. As it was shown in \cite{chen2019learning,sorkine2003high} such measures do not describe visual quality.  

The classification boundary can be regularized using IntervalNet. In Fig.~\ref{fig:missing} we present such examples. As we can see \our{} models produce single objects without empty space. To verify it, we calculate the number of connected components produced by mesh and visualize them on histograms, see Fig.~\ref{fig:hist}.  Our model produces better models for classes: Airplane, Rifle, Car. On the other hand, in the Tables, we have a similar result.  

\section{Conclusions}

In this work we introduce a new implicit field representation of 3D models. Contrary to the existing solutions, such as IM-NET, it is more light-weight and faster to train thanks to the hypernetwork architecture, while offering competitive or superior quantitative results. Finally, our method allows to incorporate interval arithmetic which enables processing entire 3D voxels, instead of their sampled version, and hence yielding more plausible and higher quality 3D reconstructions.

%

\end{document}